\documentclass{egpubl}
\usepackage{sca2026}

\CGFccby

\usepackage[T1]{fontenc}
\usepackage{dfadobe}  

\biberVersion
\BibtexOrBiblatex
\usepackage[backend=biber,bibstyle=EG,citestyle=alphabetic,backref=true]{biblatex} 
\electronicVersion
\PrintedOrElectronic

\ifpdf \usepackage[pdftex]{graphicx} \pdfcompresslevel=9
\else \usepackage[dvips]{graphicx} \fi

\usepackage{egweblnk} 

\usepackage{amsmath,amssymb,amsfonts}
\usepackage{mathtools}
\usepackage{tabularx}
\usepackage{appendix}

\usepackage{colortbl}
\usepackage{capt-of}
\usepackage{multirow}
\usepackage{makecell}
\usepackage{pgfplots}
\usepackage{pgfplotstable}
\usepackage{tikz}
\usetikzlibrary{arrows.meta,positioning}
\usepgfplotslibrary{polar}
\usepgfplotslibrary{groupplots}
\usetikzlibrary{calc}
\pgfplotsset{compat=1.18}
\usepackage{pifont}

\newcommand{\model}{IK-GAT}

\newcolumntype{Y}{>{\centering\arraybackslash}X}

\title{Amortized Inverse Kinematics via Graph Attention for Real-Time Human Avatar Animation}

\author[Khan et al.]
{\parbox{\textwidth}
    {\centering 
        Muhammad Saif Ullah Khan$^{1,2}$,
        Chen-Yu Wang$^{1}$,
        Tim Prokosch$^{2}$,
        Michael Lorenz$^{1}$,
        Bertram Taetz$^{3}$, and
        Didier Stricker$^{1,2}$ 
    }
    \\
    {\parbox{\textwidth}
        {\centering
            $^1$German Research Center for Artificial Intelligence (DFKI)  \\
            $^2$RPTU Kaiserslautern-Landau \\
            $^3$International University of Applied Sciences
        }
    }
}

\begin{document}

\teaser{
    \includegraphics[width=\linewidth]{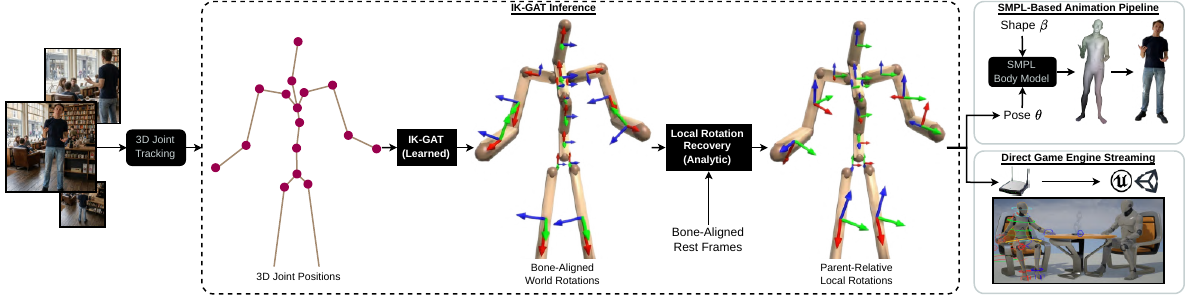}
    \centering
    \caption{\textbf{Amortized inverse kinematics.} Given 3D joint positions, IK-GAT predicts per-joint rotations in bone-aligned world space and analytically recovers standard local rotations, enabling direct animation or use in body models.}
    \label{fig:teaser-image}
}

\maketitle

\begin{abstract}
   Inverse kinematics (IK) is a core operation in animation, robotics, and
biomechanics: given Cartesian constraints, recover joint rotations under a known
kinematic tree. In many real-time human avatar pipelines, the available signal
per frame is a sparse set of tracked 3D joint positions, whereas animation
systems require \emph{joint orientations} to drive skinning. Recovering full orientations from positions is underconstrained, most
notably because twist about bone axes is ambiguous, and classical IK solvers
typically rely on iterative optimization that can be slow and sensitive to
noisy inputs. We introduce \textbf{\model}, a lightweight graph-attention
network that reconstructs full-body joint orientations from 3D joint positions
in a single forward pass. The model performs message passing over the skeletal
parent-child graph to exploit kinematic structure during rotation inference.
To simplify learning, \model~predicts rotations in a \emph{bone-aligned
world-frame} representation anchored to rest-pose bone frames. This
parameterization makes the twist axis explicit and is \emph{exactly invertible}
to standard parent-relative local rotations given the kinematic tree and rest
pose. The network uses a continuous 6D rotation representation and is trained
with a geodesic loss on $\mathrm{SO}(3)$ together with an optional
forward-kinematics consistency regularizer. \model~produces animation-ready local rotations that can directly drive a rigged
avatar or be converted to pose parameters of SMPL-like body models for real-time and online applications. With 374K parameters and over 650 FPS on CPU, \model~outperforms VPoser-based per-frame iterative optimization without warm-start at significantly lower cost, and is robust to initial pose and input noise.


\end{abstract}

\section{Introduction}
\label{sec:introduction}

Human avatar animation in production systems, such as films, gaming and VR, often uses an offline pipeline where motion capture and animation stages are distinct and asynchronous~\cite{menache2000understanding}. Motion data is first captured—traditionally using marker-based systems~\cite{geng2003reuse}—and later refined through post-processing stages such as retargeting~\cite{gleicher1998retargetting}, filtering, and manual adjustment before being applied to animated avatars. These steps ensure that the captured motion translates into visually coherent joint rotations suitable for skinning and deformation~\cite{kavan2007dqs,rumman2016skinning}. In recent years, however, motion capture has increasingly relied on sparse and markerless sensing modalities, including vision-based pose estimators and VR/AR tracking devices~\cite{ponton2023sparseposer,vanwouwe2024diffusionposer}. Such systems typically provide only 3D joint positions rather than full skeletal orientations~\cite{yi2021hierarik}, yet high-quality animation still requires \emph{joint rotations}: rotations determine skinning, preserve limb twist, and control visually salient deformations in shoulders, forearms, and thighs~\cite{kavan2007dqs,lee2013enhanced,rumman2016skinning}. Recovering these orientations from positions is fundamentally underconstrained—multiple rotation configurations can yield nearly identical joint locations, with twist about bone-aligned axes being particularly ambiguous~\cite{li2021hybrik,gildea2022kinepose,ponton2023sparseposer}.

Classical inverse-kinematics (IK) solvers address this ambiguity by introducing additional structure and constraints. Rather than inferring rotations purely from joint positions, they formulate the problem as a constrained optimization over the kinematic chain, incorporating anatomical joint limits, preferred postures, damping terms, and task-specific objectives~\cite{buss2004ik,gildea2022kinepose}. Common methods include Jacobian-based solvers, cyclic coordinate descent, and FABRIK~\cite{buss2004ik,wang1991ccd,aristidou2011fabrik}, which iteratively adjust joint rotations so that the resulting forward kinematics match the observed joint positions. While effective in controlled settings, such optimization-based methods present practical drawbacks for real-time full-body use: per-frame iteration increases latency and can amplify positional noise into rotational jitter; convergence depends on carefully tuned step sizes and damping parameters; and incorporating many joint constraints often requires manual specification and domain-specific heuristics~\cite{buss2004ik,aristidou2011fabrik,gildea2022kinepose,jiang2024manikin}. When the objective is \emph{full-body} orientation recovery~\cite{li2021hybrik,li2023niki} rather than end-effector placement, the solver must reconcile a large set of noisy positional constraints with kinematic structure and plausible articulation, making the optimization expensive and brittle~\cite{gildea2022kinepose}.

In this work, we instead learn an \emph{amortized inverse-kinematics mapping} that directly predicts joint orientations from 3D joint positions in a single forward pass. As illustrated in Fig.~\ref{fig:teaser-image}, the proposed \model~model operates as a lightweight inference module between upstream
tracking systems and downstream animation pipelines. Given tracked 3D joint positions, the network predicts per-joint rotations in a continuous 6D representation~\cite{zhou2019continuity}. These predictions are converted to $\mathrm{SO}(3)$ using Gram--Schmidt orthogonalization~\cite{zhou2019continuity}, yielding canonical \emph{bone-aligned world rotations} whose primary axis follows the bone direction. Using precomputed rest-pose bone frames, an analytic mapping then recovers rig-specific local rotations required by standard animation rigs.

This decomposition separates learned and deterministic components of the problem. The network focuses on predicting a geometrically consistent world-frame orientation for each bone, while kinematic composition and localization are recovered exactly through the known skeletal structure. The
resulting system acts as a lightweight neural inverse-kinematics module~\cite{bensadoun2022neuralik} that can be integrated into multiple animation pipelines. For example, predicted joint orientations can drive SMPL-like~\cite{loper2015smpl} parametric body models within existing workflows, or they can be streamed directly to a target character rig or game engine without relying on an intermediate body model, enabling real-time animation from joint positions only.

\noindent
\textbf{Contributions.}
This paper makes the following contributions:
\begin{itemize}
    \item We introduce \textbf{\model}, a lightweight graph-attention network that predicts physically-plausible per-joint orientations from 3D joint positions at high framerates, enabling real-time animation.
    \item We propose using \textbf{bone-aligned rotations in world coordinates} for a more stable learning objective while remaining \emph{exactly invertible} to rig-specific world-frames and standard parent-relative local rotations given a known rest pose and a kinematic tree.
    \item We incorporate an explicit \textbf{kinematic prior} by using the skeletal parent--child graph as the message-passing topology, enabling stable and efficient learning of articulated motion.
    \item We evaluate across heterogeneous motion sources for SMPL and Unreal Engine (UE) avatar animation, and analyze robustness under noisy joint inputs.
\end{itemize}

\section{Related Work}
\label{sec:related_work}

We review prior work on IK, optimization-based body model
fitting, learning-based IK, and skeleton-aware neural architectures, and position our setting relative to Human Mesh Recovery (HMR).

\noindent\textbf{Inverse kinematics in robotics and animation.}
Inverse kinematics seeks joint parameters that satisfy Cartesian constraints under a known kinematic chain. Closed-form analytic solutions exist only for restricted mechanisms with
special structure, such as those satisfying Pieper's conditions~\cite{pieper1968}. For general articulated figures, most systems rely on iterative numerical optimization. Common approaches include Jacobian-based solvers~\cite{buss2004ik}, cyclic coordinate descent (CCD)~\cite{wang1991ccd}, and FABRIK~\cite{aristidou2011fabrik}. In biomechanics analysis, OpenSim~\cite{delp2007opensim} formulates IK as a constrained least-squares problem that matches observed markers or joints, but this only works reliably when enough markers to recover an observable orientation per segment or joint are available. This is different to IK from joints only which usually does not allow for observable internal rotation estimation. While widely used, iterative solvers are sensitive to observation noise and solver hyperparameters, and their per-frame optimization cost can become prohibitive when many joints and constraints must be satisfied in real time~\cite{aristidou2009inverse}.

\noindent\textbf{Optimization-based body model fitting.}
Parametric human body models such as SMPL and SMPL-X provide a standardized pose representation and differentiable forward kinematics for joints and mesh vertices~\cite{loper2015smpl,pavlakos2019smplx}. Optimization-based fitting methods such as SMPLify recover pose parameters from 2D or 3D observations using priors and regularization
\cite{bogo2016smplify}. Learned pose priors (e.g., VPoser) further constrain solutions toward plausible human configurations~\cite{ghorbani2019vposer}. Although effective for offline reconstruction, these approaches remain per-instance optimization procedures and can introduce latency and temporal instability when applied to real-time animation.

\noindent\textbf{Human mesh recovery.}
Human Mesh Recovery (HMR) methods regress body pose and shape directly from images or keypoints. Early work introduced end-to-end regression of SMPL parameters from monocular images~\cite{kanazawa2018hmr}, and later systems improved robustness and accuracy through stronger architectures and temporal reasoning~\cite{goel2023humans4d,baradel2024multihmr}.
Unlike HMR, our setting assumes that 3D joint positions are already available from an upstream tracking system and focuses exclusively on recovering stable joint orientations suitable for animation pipelines.

\noindent\textbf{Learning-based and hybrid inverse kinematics.}
Learning-based IK amortizes the mapping from Cartesian constraints to joint parameters using neural networks. Several works combine learned inference with analytic structure. HybrIK recovers rotations from RGB images using a hybrid analytic--neural formulation with explicit twist handling~\cite{li2021hybrik}. HierarIK exploits the kinematic hierarchy to improve robustness under noisy keypoints~\cite{yi2021hierarik}, while MANIKIN and related systems target sparse tracking scenarios and incorporate biomechanical constraints to produce plausible full-body motion~\cite{jiang2024manikin}. Other sparse motion tracking methods like Deep Inertial Poser~\cite{huang2018deep} and TransPose~\cite{yi2021transpose} further demonstrate the feasibility of learning-based IK under limited observations and using IMU data. Our work follows this direction but focuses specifically on only dense 3D joint inputs and real-time single-frame rotation recovery for character animation.

\noindent\textbf{Skeleton-aware learning.}
Human skeletons naturally form graphs defined by anatomical connectivity. Graph neural networks have therefore become a common architecture for skeleton-based reasoning. ST-GCN introduced spatio-temporal graph convolutions for skeleton sequences~\cite{yan2018stgcn}, while SemGCN learned semantic edge weights for pose regression~\cite{zhao2019semgcn}. Graph Attention Networks further introduced attention-based message passing over graph neighborhoods~\cite{velickovic2018gat}. These architectures demonstrate the benefit of explicitly modeling the kinematic tree structure, motivating our use of graph attention over skeletal connectivity.

\noindent\textbf{Rotation representations.}
Learning rotations requires representations compatible with Euclidean neural networks while respecting $\mathrm{SO}(3)$ geometry. The continuous 6D representation predicts two rotation-matrix columns followed by orthonormalization, avoiding discontinuities present in Euler angles and quaternions~\cite{zhou2019continuity}. Rotation errors are commonly measured using geodesic distances on $\mathrm{SO}(3)$~\cite{huynh2009metrics}. Our model adopts these standard choices while introducing a bone-aligned world-frame parameterization.
\section{Method}
\label{sec:method}

\subsection{Problem formulation}
\label{sec:problem}

We consider a temporally causal stream of 3D joint positions expressed in root-centered coordinates (root space), observed at time $t$ as a rooted kinematic tree with pose 
\begin{equation}
\tilde{\mathbf{J}}_t = \left\{ \tilde{\mathbf{j}}_t^{(i)} \in \mathbb{R}^3 \right\}_{i=0}^{N-1},
\end{equation}
where $\tilde{\mathbf{j}}^{(0)}_t = \mathbf{0}$ denotes the root joint, $N$ is the number of joints, $i$ is the joint index, and joints are sorted in a parent-first order. Let $\pi(i)\in[-1,N-1]$ denote the parent of joint $i$, with $\pi(0)=-1$ for the root. The $k$-th ancestor of joint $i$ is defined recursively as $$\pi^{0}(i) = i, \qquad \pi^{k}(i) = \pi\!\big(\pi^{k-1}(i)\big), \quad k \ge 1,$$ and for each joint $i$, there exists a finite depth $d(i)=K$ such that $\pi^{K}(i)=0$ and $\pi^{K+1}(i)=-1$; equivalently, repeated application of $\pi$ reaches the root in finitely many steps.

To support real-time, online, and memory-efficient avatar animation, we aim to recover parent-relative local joint rotations suitable for standard animation pipelines, in a single per-frame forward pass of a lightweight neural network without iterative optimizations, access to future poses, or an unbounded access to past poses.

\subsubsection{Forward kinematics (FK)}
Let $\mathbf{R}^{(i)}_{t,\mathrm{loc}} \in \mathrm{SO}(3)$ denote the parent-relative local rotation of joint $i$ at time $t$, and define $\mathbf{R}^{(0)}_{t,\mathrm{w}}=\mathbf{R}^{(0)}_{t,\mathrm{loc}}$ as the world rotation of the root. World-space joint rotation matrices are then obtained recursively as
\begin{align}
    \mathbf{R}^{(i)}_{t,\mathrm{w}} &=
    \begin{cases}
        \mathbf{R}^{(0)}_{t,\mathrm{loc}}, & i=0, \\[1ex]
        \mathbf{R}^{(\pi(i))}_{t,\mathrm{w}} \mathbf{R}^{(i)}_{t,\mathrm{loc}}   & \text{otherwise}.
    \end{cases}
\label{eq:world_rotations}
\end{align}
Let $\bar{\mathbf{J}} = \left\{ \bar{\mathbf{j}}^{(i)} \in \mathbb{R}^3 \right\}_{i=0}^{N-1}$ be the unposed, template rig with joints at rest, $\bar{\mathbf{d}}^{(i)}=\bar{\mathbf{j}}^{(i)}-\bar{\mathbf{j}}^{(\pi(i))}$ be the fixed rest-pose offset from the parent to joint $i \in (0, N)$, and $\bar{\mathbf{d}}^{(0)}=\mathbf{0}$ for the root, and $\bar{\mathbf{d}}^{(i)}\neq\mathbf{0}$ for $i > 0$. Root-space joint positions are defined recursively as
\begin{equation}
\tilde{\mathbf{j}}^{(i)}_t = 
\begin{cases}
    \mathbf{0}, & i=0, \\[1ex]
    \tilde{\mathbf{j}}^{(\pi(i))}_t + \mathbf{R}^{(\pi(i))}_{t,\mathrm{w}} \, \bar{\mathbf{d}}^{(i)}, & \text{otherwise}.    
\end{cases}
\label{eq:world_positions}
\end{equation}
We denote this mapping by $\mathrm{FK}\big(\{\mathbf{R}^{(i)}_{t,\mathrm{loc}}\}_{i=0}^{N-1}\big)$.

\subsection{Inverse kinematics via bone-aligned world frames}
\label{sec:bone-frames}
\begin{figure}[t]
    \centering
    \includegraphics[width=0.5\linewidth]{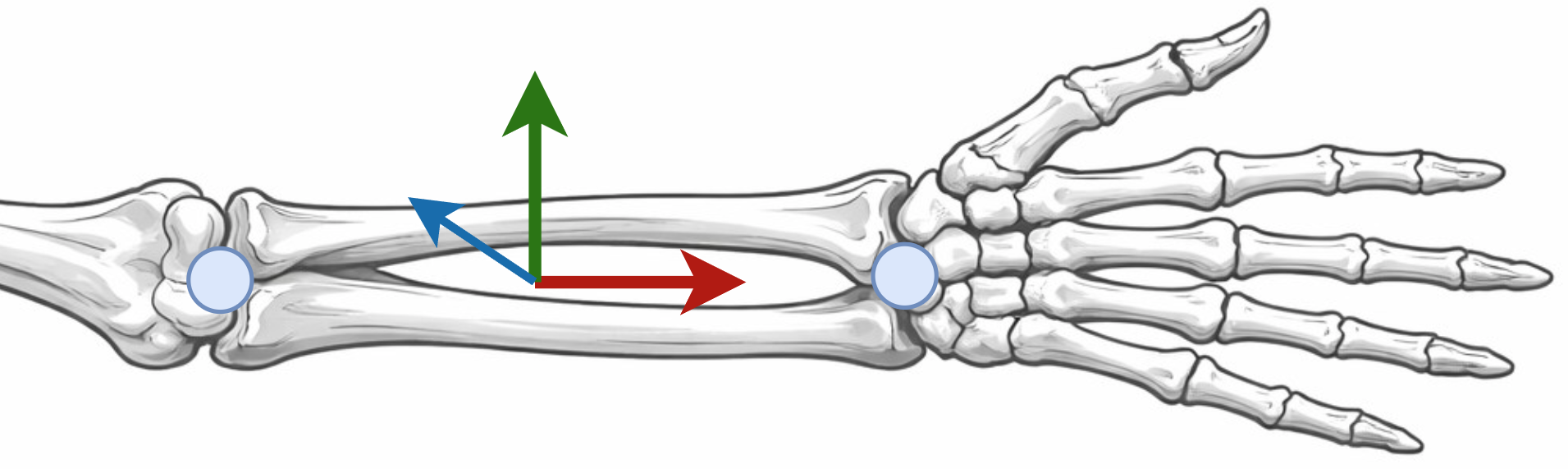}
    \includegraphics[width=0.5\linewidth]{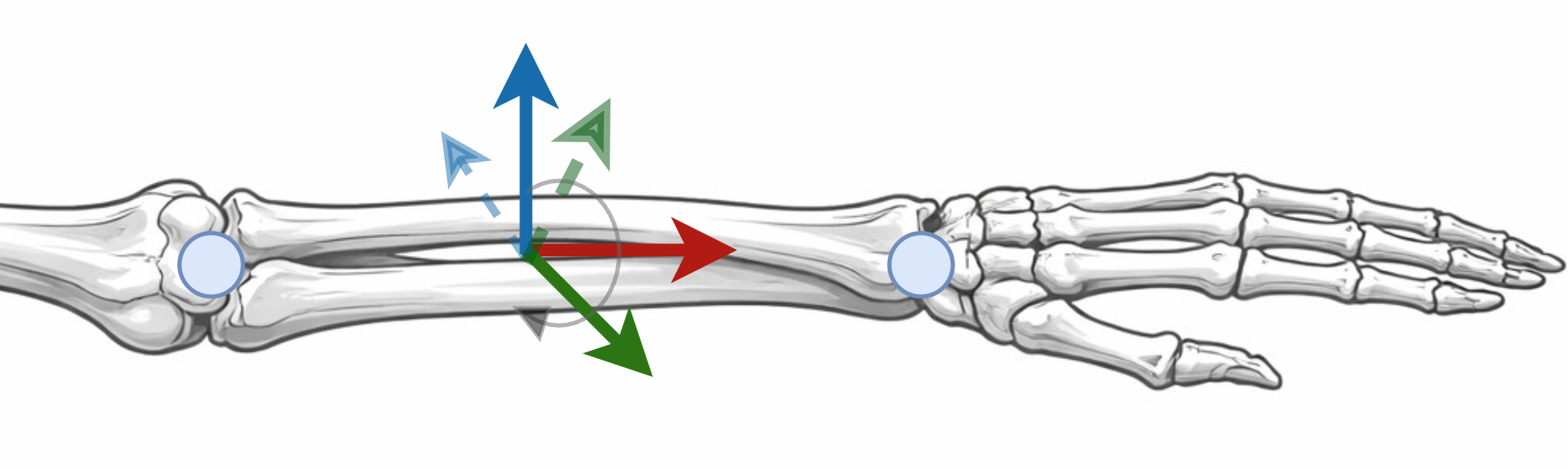}
    \caption{\textbf{Twist ambiguity problem.} Identical joint positions (blue dots) are produced by any rotation around the bone axis (red). This makes the inverse problem ill-posed without strong anatomical priors. Neural IK methods attempt to learn these priors from data.}
    \label{fig:bone-frames-vis}
\end{figure}

Inverse kinematics (IK) is the inverse task of recovering a set of local rotations given 3D joints, such that $\mathrm{FK}(\{\mathbf{R}^{(i)}_{t,\mathrm{loc}}\}_{i=0}^{N-1}) \approx \tilde{\mathbf{J}}_t$.
This is ill-posed: joint positions underconstrain rotations, most notably
through twist around bone-aligned axes and in the presence of noisy or missing
joints, as shown in Fig.~\ref{fig:bone-frames-vis}.

\noindent
\textbf{Key idea.}
We separate the IK problem into a learned component and a deterministic component. Rather than regressing local rotations directly, we learn an intermediate representation that (i) is stable for regression and (ii) admits an exact, closed-form conversion to standard local rotations. The network predicts per-joint rotation matrices representing \emph{bone-aligned world frames}, while the conversion to animation-compatible parent-relative local rotations is analytic and lossless. This reduces the complexity of what must be learned: the model is not required to learn kinematic composition rules or rest-pose basis conventions, only the pose-dependent orientations in a canonicalized frame.

The bone-aligned world rotation matrices differ from the standard world-space rotation matrices in how the underlying local coordinate frames are defined. In typical rigged avatars, each joint is associated with an author-defined local frame whose orientation relative to the bone is arbitrary and varies across rigs. Consequently, the corresponding world rotation matrices depend on these rig-specific conventions. In contrast, we define a canonical local frame at each joint directly from the rest-pose geometry. This yields a consistent, bone-aligned frame whose $x$-axis follows the bone direction and whose remaining axes are fixed deterministically.

\subsubsection{Rest-pose bone frames (precomputed once per rig)}
Let $\mathbf{u}$ be a fixed unit vector denoting the global up direction. For each joint $i$ in a given rig with rest-pose $\bar{\mathbf{J}}$, we select a \emph{primary child} $c(i)$ using a deterministic choice among children. For joints with multiple children, we prefer the child whose offset aligns most with global up, and break ties by longer bone length.

We define a directed edge $(s(i), t(i))$ for each joint $i$ as
\begin{equation}
(s(i), t(i)) =
\begin{cases}
(i,\, c(i)), & \text{if } i \text{ has at least one child},\\
(\pi(i),\, i), & \text{otherwise},
\end{cases}
\end{equation}
where $\pi(i)$ denotes the parent of $i$. Assuming nonzero bone lengths, the bone direction at rest is then defined as
\begin{equation}
\bar{\mathbf{x}}^{(i)} =
\frac{\bar{\mathbf{j}}^{(t(i))}-\bar{\mathbf{j}}^{(s(i))}}
{\left\|\bar{\mathbf{j}}^{(t(i))}-\bar{\mathbf{j}}^{(s(i))}\right\|_2}.
\label{eq:rest_bone_axis}
\end{equation}
To fix the remaining degree of freedom (twist), we choose a reference direction
\begin{equation}
\mathbf{r}^{(i)} =
\begin{cases}
\mathbf{u}, & i=0,\\
\bar{\mathbf{y}}^{(\pi(i))}, & \text{otherwise},
\end{cases}
\label{eq:rest_ref_axis}
\end{equation}
where $\bar{\mathbf{y}}^{(\pi(i))}$ is the parent’s propagated reference $y$-axis, recursively computed as defined in \eqref{eq:ortho}. We then construct an orthonormal, right-handed basis via Gram-Schmidt \cite{zhou2019continuity}:
\begin{align}
\label{eq:ortho}
\tilde{\mathbf{y}}^{(i)} &=
\mathbf{r}^{(i)} - \left\langle \mathbf{r}^{(i)}, \bar{\mathbf{x}}^{(i)} \right\rangle\bar{\mathbf{x}}^{(i)}, \qquad
\bar{\mathbf{y}}^{(i)} =
\frac{\tilde{\mathbf{y}}^{(i)}}{\|\tilde{\mathbf{y}}^{(i)}\|_2}, \\
\bar{\mathbf{z}}^{(i)} &= \bar{\mathbf{x}}^{(i)} \times \bar{\mathbf{y}}^{(i)}.
\end{align}
In the degenerate case where $\mathbf{r}^{(i)}$ is (near-)collinear with $\bar{\mathbf{x}}^{(i)}$, the projection in \eqref{eq:ortho} yields $\|\tilde{\mathbf{y}}^{(i)}\|_2 \approx 0$, making the normalization ill-defined. In this case, we detect the degeneracy and reapply the construction using a fallback reference direction $\mathbf{r}_{\mathrm{fb}}$, chosen as the global up vector $\mathbf{u}$. The resulting frame is well-defined provided at least one of $\mathbf{r}^{(i)}$ or $\mathbf{r}_{\mathrm{fb}}$ is not collinear with $\bar{\mathbf{x}}^{(i)}$.

In practice, this condition is easily satisfied for human-like kinematic trees in a rest pose: bone directions are diverse and rarely align with a single global axis across parent-child chains, ensuring that either the propagated parent reference or the global up vector provides a valid, non-collinear direction.

The resulting \emph{bone-aligned rest frame} is represented by
\begin{equation}
\mathbf{B}^{(i)}_{\mathrm{rest}} =
\big[\bar{\mathbf{x}}^{(i)}\;\; \bar{\mathbf{y}}^{(i)}\;\; \bar{\mathbf{z}}^{(i)}\big]
\in \mathrm{SO}(3),
\label{eq:rest_frame}
\end{equation}
which is the rotation matrix mapping the bone-local rest frame to world coordinates at rest. This computation is performed once per target rig and does not depend on the observed motion.

\subsubsection{Bone-aligned world frame representations}
Let $\mathbf{R}^{(i)}_{t,\mathrm{w}}$ be the joint world rotation under FK.
We define the rotation matrix representing the \emph{bone-aligned world frame}
\begin{equation}
\mathbf{R}^{(i)}_{t,\mathrm{bone}} =
\mathbf{R}^{(i)}_{t,\mathrm{w}} \, \mathbf{B}^{(i)}_{\mathrm{rest}}.
\label{eq:bone_world}
\end{equation}
Intuitively, the matrix $\mathbf{R}^{(i)}_{t,\mathrm{bone}}$ maps the canonical bone-aligned local frame to world coordinates. By construction, the local $x$-axis of this frame is defined from the rest-pose bone direction, and under $\mathbf{R}^{(i)}_{t,\mathrm{bone}}$ it is transformed to the current world-space direction of the rest-defined local $x$-axis. This makes the twist axis (rotation about the bone direction) explicit while preserving all three rotational DoFs.


\subsubsection{Exact invertibility to parent-relative local rotations}
Given $\mathbf{R}^{(i)}_{t,\mathrm{bone}}$ and $\mathbf{B}^{(i)}_{\mathrm{rest}}$, we recover world rotations exactly via
\begin{equation}
\mathbf{R}^{(i)}_{t,\mathrm{w}} =
\mathbf{R}^{(i)}_{t,\mathrm{bone}}\left(\mathbf{B}^{(i)}_{\mathrm{rest}}\right)^\top,
\label{eq:recover_world}
\end{equation}
and then recover local rotations by the standard parent-relative relation
\begin{equation}
\mathbf{R}^{(i)}_{t,\mathrm{loc}}=
\begin{cases}
\mathbf{R}^{(i)}_{t,\mathrm{w}}, & \pi(i)=-1,\\
\left(\mathbf{R}^{(\pi(i))}_{t,\mathrm{w}}\right)^\top \mathbf{R}^{(i)}_{t,\mathrm{w}}, & \text{otherwise}.
\end{cases}
\label{eq:recover_local}
\end{equation}
\eqref{eq:recover_world} follows from orthogonality of $\mathbf{B}^{(i)}_{\mathrm{rest}}$, and \eqref{eq:recover_local} is the standard parent-relative composition identity. Empirically, across 0.98M 22-joint data frames, local rotation recovery $\phi(\cdot)$ is numerically exact up to $\texttt{float32}$ precision, with maximum matrix-space error $\max_t |\mathbf{R}_{t,\mathrm{loc}}-\phi(\mathbf{R}_{t,\mathrm{bone}})|_F = 1.8\mathrm{e}{-5};$ (mean $1.0\mathrm{e}{-5}$).

\begin{figure*}[t]
    \includegraphics[width=\linewidth]{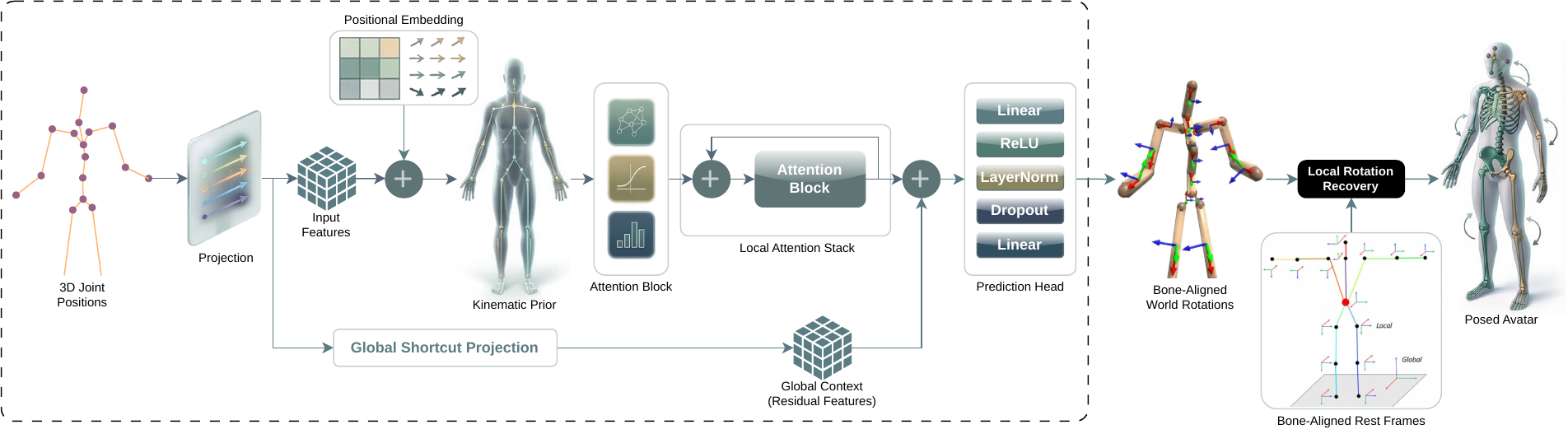}
    \caption{\textbf{\model~architecture.} The model accepts 3D joint positions as input, which are projected and augmented with learnable positional embeddings. The core encoder leverages a \textbf{Kinematic Prior}, defined by the skeletal parent-child relationships, to structure the Graph Attention (GAT) layers. The network employs a dual-residual design, featuring both local skip connections within GAT blocks and a global shortcut from the input projection. The head regresses rotation matrices representing bone-aligned world frames via a 6D parameterization.}
    \label{fig:model-architecture}
\end{figure*}

\subsection{\model: Graph-attention inverse kinematics}
\label{sec:model-architecture}

Figure~\ref{fig:model-architecture} summarizes the proposed \model~architecture. We learn a mapping from root-space joint positions to rotation matrices representing bone-aligned world frames:
\begin{equation}
\tilde{\mathbf{J}}_t \;\mapsto\; \left\{\hat{\mathbf{R}}^{(i)}_{t,\mathrm{bone}}\right\}_{i=0}^{N-1}.
\end{equation}

\subsubsection{Skeletal graph (kinematic prior)}
The kinematic tree can also be denoted by a graph \(\mathcal{G}=(\mathcal{V},\mathcal{E})\) with one node per joint and edges connecting each joint to its parent. We use the kinematic tree as an undirected message-passing graph by adding bidirectional edges \((i,\pi(i))\) and \((\pi(i),i)\) for each non-root joint. This imposes an inductive bias that information flows along anatomically meaningful connections.

\subsubsection{Node features and embedding}
\label{sec:node-feature-embedding}
Each joint is represented by a feature vector comprising its observed root-space position
\(
\mathbf{x}^{(i)} = \tilde{\mathbf{j}}^{(i)}_t \in \mathbb{R}^3,
\) optionally concatenated with additional features. We then project the feature vector to a latent dimension \(F\) and add a learnable joint-index embedding:
\begin{equation}
\mathbf{h}^{(i)}_0 = \mathbf{W}_{\mathrm{in}}\mathbf{x}^{(i)} + \mathbf{e}^{(i)}.
\end{equation}
This allows the network to specialize behavior per anatomical joint even under symmetric or ambiguous positional patterns.

\subsubsection{Graph-attention blocks with distal refinement}
We apply $D$ stacked multi-head Graph Attention layers \cite{velickovic2018gat}
on the kinematic graph. In the implementation, each GAT layer operates on the
bidirectional parent-child graph and also includes self-loops, so each joint
aggregates both its own feature and messages from its kinematic neighbors.
For layer $l$,
\begin{equation}
\tilde{\mathbf{h}}^{(i)}_{l+1}
=
\mathrm{LN}\!\left(
\mathrm{ELU}\!\left(
\sum_{j\in\mathcal{N}^{+}(i)}
\alpha^{(l)}_{ij}\,\mathbf{W}^{(l)}\mathbf{h}^{(j)}_{l}
\right)\right),
\end{equation}
where $\mathcal{N}^{+}(i)=\mathcal{N}(i)\cup\{i\}$ denotes the neighbor set
augmented with the self-loop, $\alpha^{(l)}_{ij}$ are learned attention weights,
and $\mathrm{LN}(\cdot)$ is LayerNorm.

In addition, we use a \emph{global} shortcut that bypasses the entire GAT stack.
A learned linear projection of the raw per-joint input is added to the final
pre-head features:
\begin{equation}
\mathbf{r}^{(i)} = \mathbf{W}_{\mathrm{skip}}\mathbf{x}^{(i)}, \qquad
\mathbf{z}^{(i)} = \mathbf{h}^{(i)}_{D} + \mathbf{r}^{(i)}.
\end{equation}
This shortcut preserves direct access to the observed geometric input while the
graph-attention layers model inter-joint dependencies.

When enabled, the model also uses a \emph{local kinematic refinement} branch in
the later half of the GAT stack rather than a standard hidden-state residual.
Let $\mathcal{C}(i)$ denote the children of joint $i$, and define the immediate
kinematic neighborhood
\begin{equation}
\mathcal{K}(i) =
\bigl(\{\pi(i)\}\ \text{if } \pi(i)\neq -1 \bigr)\ \cup\ \mathcal{C}(i).
\end{equation}
From previous-layer features, we compute the neighborhood mean
\begin{equation}
\bar{\mathbf{h}}^{(i)}_{l}
=
\frac{1}{|\mathcal{K}(i)|}
\sum_{j\in\mathcal{K}(i)} \mathbf{h}^{(j)}_{l},
\end{equation}
and form a learned local correction
\begin{equation}
\Delta\mathbf{h}^{(i)}_{l}
=
\mathbf{W}_{\mathrm{loc}}
\left(\bar{\mathbf{h}}^{(i)}_{l} - \mathbf{h}^{(i)}_{l}\right).
\end{equation}
This correction is applied only for layers in the later half of the stack and
only on a distal-joint set $\mathcal{D}$ consisting of end effectors and their
immediate parents:
\begin{equation}
\mathbf{h}^{(i)}_{l+1}
=
\tilde{\mathbf{h}}^{(i)}_{l+1}
+
\mathbb{1}[i\in\mathcal{D}]\,\Delta\mathbf{h}^{(i)}_{l}.
\end{equation}

Thus, instead of a generic residual connection between consecutive hidden states, this is a structured, kinematic local refinement term that selectively adjusts distal chains such as wrist--elbow, ankle--foot, and neck--head, while proximal joints are handled primarily by graph message passing and the global shortcut.

\subsubsection{Rotation head: 6D parameterization}
Each joint feature is mapped to a 6D rotation representation
$\hat{\mathbf{r}}^{(i)}\in\mathbb{R}^6$, which is converted to a valid rotation
matrix via Gram-Schmidt. Writing $\hat{\mathbf{r}}^{(i)}=[\mathbf{a}_1^\top\;\mathbf{a}_2^\top]^\top$ with
$\mathbf{a}_1,\mathbf{a}_2\in\mathbb{R}^3$:
\begin{align}
\hat{\mathbf{x}}^{(i)} &= \frac{\mathbf{a}_1}{\|\mathbf{a}_1\|_2}, \\
\tilde{\mathbf{y}}^{(i)} &=
\mathbf{a}_2 - \left\langle \mathbf{a}_2, \hat{\mathbf{x}}^{(i)} \right\rangle \hat{\mathbf{x}}^{(i)}, \qquad
\hat{\mathbf{y}}^{(i)} = \frac{\tilde{\mathbf{y}}^{(i)}}{\|\tilde{\mathbf{y}}^{(i)}\|_2}, \\
\hat{\mathbf{z}}^{(i)} &= \hat{\mathbf{x}}^{(i)} \times \hat{\mathbf{y}}^{(i)},
\end{align}
and $\hat{\mathbf{R}}^{(i)}_{t,\mathrm{bone}} =
[\hat{\mathbf{x}}^{(i)}\;\hat{\mathbf{y}}^{(i)}\;\hat{\mathbf{z}}^{(i)}]\in\mathrm{SO}(3)$.

\subsection{Training objectives}
\label{sec:losses}

\noindent
\textbf{Geodesic rotation loss.}
We supervise predicted rotation matrices using the geodesic distance on $\mathrm{SO}(3)$ \cite{huynh2009metrics}:
\begin{align}
d_{\mathrm{geo}}(A,B) &= \arccos\!\left( \frac{\mathrm{tr}(A^\top B)-1}{2} \right),
\end{align}
where $A,B\in\mathrm{SO}(3)$, and
\begin{align}
\mathcal{L}_{\mathrm{rot}} &=
\frac{1}{N}\sum_{i=0}^{N-1}
d_{\mathrm{geo}}\!\left(\hat{\mathbf{R}}^{(i)}_{t,\mathrm{bone}}, \mathbf{R}^{(i)}_{t,\mathrm{bone}}\right).
\label{eq:geo_metric}
\end{align}
Since $-1 \leq \mathrm{tr}(U) \leq 3$ for any $U \in \mathrm{SO}(3)$, \eqref{eq:geo_metric} is well-defined.

\noindent
\textbf{Forward-kinematics consistency.}
We optionally enforce that the predicted rotations reproduce the input root-space joints after analytic recovery of predicted world and local rotations (Eq.~\ref{eq:recover_world}--\ref{eq:recover_local}) and forward kinematics (Eq.~\ref{eq:world_rotations}--\ref{eq:world_positions}). Let
\begin{equation}
\hat{\mathbf{J}}^{\mathrm{rec}}_t =
\left\{ \hat{\mathbf{j}}^{(i),\mathrm{rec}}_t \right\}_{i=0}^{N-1}
=
\mathrm{FK}\big(\{\hat{\mathbf{R}}^{(i)}_{t,\mathrm{loc}}\}_{i=0}^{N-1}\big)
\end{equation}
denote the root-space joints reconstructed from the predicted local rotations. We then define
\begin{equation}
\mathcal{L}_{\mathrm{fk}} =
\frac{1}{N}\sum_{i=0}^{N-1}\left\|\hat{\mathbf{j}}^{(i),\mathrm{rec}}_t-\tilde{\mathbf{j}}^{(i)}_t\right\|_2^2.
\end{equation}
The total loss is
\begin{equation}
\mathcal{L} = \mathcal{L}_{\mathrm{rot}} + \alpha\mathcal{L}_{\mathrm{fk}},
\end{equation} where $\alpha$ is a free hyperparameter controlling the influence of the FK consistency regularization.

\subsection{Inference and deployment}
\label{sec:inference}

At inference time, the network predicts \(\hat{\mathbf{R}}^{(i)}_{t,\mathrm{bone}}\) from the root-space input \(\tilde{\mathbf{J}}_t\) in a single pass. We recover predicted world rotation matrices via \eqref{eq:recover_world} and then predicted parent-relative local rotation matrices via \eqref{eq:recover_local}. Because the network operates in root space, the global root translation \(\mathbf{t}\) is not predicted by the model and is added separately when world-space placement is required.

Given a global root translation $\mathbf{t}$, world-space joint positions can be obtained as
\begin{equation}
\mathbf{J}_t = \left\{ \tilde{\mathbf{j}}^{(i)}_t + \mathbf{t} \right\}_{i=0}^{N-1}.
\end{equation}
Since the IK mapping depends only on relative joint configuration, we operate entirely in root space during learning and inference.

The only rig-specific inputs required are the kinematic tree and the rest-frame rotation matrices \(\{\mathbf{B}^{(i)}_{\mathrm{rest}}\}\), computed once per rig. The resulting local rotations can be applied directly to a rigged avatar, or converted to axis--angle parameters for parametric body models such as SMPL \cite{loper2015smpl}. In that case, the recovered root rotation \(\hat{\mathbf{R}}^{(0)}_{t,\mathrm{loc}}=\hat{\mathbf{R}}^{(0)}_{t,\mathrm{w}}\) corresponds to the global orientation parameter, the non-root local rotations \(\{\hat{\mathbf{R}}^{(i)}_{t,\mathrm{loc}}\}_{i=1}^{N-1}\) correspond to the other pose parameters, and the root translation \(\mathbf{t}\) corresponds to the global translation parameter. Shape coefficients \(\boldsymbol{\beta}\) are not estimated by \model\ and must come from an external source (e.g., a fixed mean shape, a subject-specific estimate, or a downstream fitting step).
\section{Experiments}
\label{sec:experiments}

We evaluate \model\ as an amortized inverse-kinematics module that sits between upstream 3D joint tracking and downstream avatar animation. The experiments address four questions: whether single-pass inference can match or exceed iterative IK on a standardized full-body benchmark; whether the proposed bone-aligned world representation simplifies learning relative to direct local-rotation regression; whether anatomically structured graph attention is more effective than generic dense interaction; and whether the resulting model transfers to a production rig under noisy tracking. We therefore use two complementary benchmarks. The SMPL benchmark isolates pose recovery under standardized kinematics and provides the main quantitative comparison. The UE5 benchmark tests whether the analytically recovered local rotations remain directly useful on a production rig. Unless otherwise stated, all models operate causally on single frames of root-space joints, without future context or iterative refinement. 

\begin{table*}[ht]
\small
\setlength{\tabcolsep}{3pt}
\caption{\textbf{AMASS SMPL benchmark.} Lower is better. IK-GAT is best on most reported metrics. Relative to the strongest feed-forward baseline (Transformer), it reduces test MPJAE from 9.06 to 7.43 (\(-18.0\%\)) while using \(8.6\times\) fewer parameters (0.37M vs. 3.17M). Relative to the strongest iterative IK baseline on MPJAE (VPoser), it further reduces test MPJAE from 9.24 to 7.43 despite requiring only a single forward pass instead of 200 optimization steps. Optimization baselines fit both pose and shape from scratch for every frame, whereas feed-forward methods predict pose only and use ground-truth shape for body-model metrics.}
\label{tab:amass_results}
\begin{tabularx}{\linewidth}{|l|ccccc|ccccc|YY|}
\hline
& \multicolumn{5}{c|}{\textbf{Validation Set}} & \multicolumn{5}{c|}{\textbf{Test Set}} & \multicolumn{2}{c|}{\textbf{Runtime}} \\
\hline
\textbf{Method} & \textbf{MPJAE} & \textbf{MPJPE} & \textbf{P-MPJPE} & \textbf{MPVE} & \textbf{P-MPVE} & \textbf{MPJAE} & \textbf{MPJPE}  &\textbf{P-MPJPE}& \textbf{MPVE}    & \textbf{P-MPVE} & \textbf{Params(M)} & \textbf{MAdds(M)} \\
\hline
Lower Bound   & 22.95 & 691.03 & 160.54 & 849.55 & 211.32 & 23.88 & 650.99 & 136.47 & 835.76 & 173.71 & - & - \\
\hline
LBFGS              & 16.92 &  45.34 &  17.38 &  57.15 &  28.59 & 14.93 &  40.42 &  \textbf{11.30} &  51.75 &  30.79 & - & - \\
VPoser       & 10.25 &  50.05 &  22.70 &  63.04 & 33.30 &  9.24 &  44.13 &  23.07 &  59.57 &  38.35 & - & - \\
SMPLify3D          & 10.61 &  19.58 &   \textbf{9.56} &  34.72 & 24.99 & 11.86 &  34.61 &  14.03 &  48.91 &  31.99 & - & - \\
\hline
MLP$\ddagger$      & 45.82 & 227.23 & 136.19 & 306.74 & 151.72 & 39.32 & 211.70 & 117.86 & 299.64 & 130.95 & 2.11 & 2.23 \\
QuaterGCN$\dagger$ & 16.64 &  56.94 &  41.20 &  73.26 &  44.44 & 14.72 &  48.18 &  36.37 &  57.44 &  37.76 & 0.13 & 2.70 \\
QuaterGCN-6D$\dagger$& 15.35 &  44.26 &  34.16 &  56.28 &  37.28 & 14.31 &  37.94 &  31.78 &  44.19 &  30.51 & 0.13 & 2.71 \\
STIK$\ddagger$     & 13.83 &  50.41 &  37.95 &  57.73 &  37.93 & 12.57 &  43.05 &  34.51 &  44.93 &  31.90 & 1.32 & 174.87 \\
Transformer$\ddagger$&  9.48 &  22.86 &  17.40 &  37.61 &  27.70 &  9.06 &  22.54 &  16.53 &  33.08 &  24.55 & 3.17 & 2.23 \\
\rowcolor{gray!20}
\model~(ours)      &  \textbf{7.98} &  \textbf{18.19} &  \underline{14.21} &  \textbf{27.90} &  \textbf{21.73} &  \textbf{7.43} &  \textbf{18.60} &  \underline{12.80} &  \textbf{24.47} &  \textbf{18.08} & 0.37 & 6.10 \\
\hline
\end{tabularx}
\end{table*}


\subsection{Evaluation metrics}
\label{sec:metrics}

Our primary metric is \textbf{mean per-joint angular error (MPJAE)} in parent-relative local rotation space. This aligns the evaluation with the final animation-ready rotation representation recovered at inference time. Given predicted bone-aligned world rotations \(\hat{\mathbf{R}}^{(i)}_{t,\mathrm{bone}}\) and ground-truth bone-aligned world rotations \(\mathbf{R}^{(i)}_{t,\mathrm{bone}}\), we first recover the corresponding local rotations \(\hat{\mathbf{R}}^{(i)}_{t,\mathrm{loc}}\) and \(\mathbf{R}^{(i)}_{t,\mathrm{loc}}\) using Eq.~\ref{eq:recover_world}--\ref{eq:recover_local}. We then define
\begin{equation}
\mathrm{MPJAE}
=
\frac{180}{\pi N}
\sum_{i=0}^{N-1}
\arccos\!\left(
\frac{
\mathrm{tr}\!\left(
(\hat{\mathbf{R}}^{(i)}_{t,\mathrm{loc}})^\top
\mathbf{R}^{(i)}_{t,\mathrm{loc}}
\right)-1
}{2}
\right),
\end{equation}
reported in degrees, and averaged over all evaluation samples.

We also report \textbf{swing} and \textbf{twist} errors as secondary rotation diagnostics derived directly from the predicted and ground-truth bone-aligned world frames. Swing error is measured as the angular distance between the predicted and ground-truth first frame axes, \(\hat{\mathbf{x}}^{(i)}\) and \(\mathbf{x}^{(i)}\), while twist error is measured as the angular distance between the corresponding second frame axes, \(\hat{\mathbf{y}}^{(i)}\) and \(\mathbf{y}^{(i)}\). These diagnostics are useful because they reveal whether an improvement comes primarily from recovering bone direction or from resolving the residual in-plane ambiguity around the bone axis.

For the SMPL benchmark, we additionally report \textbf{MPJPE} and \textbf{P-MPJPE} on joints, and \textbf{MPVE} and \textbf{P-MPVE} on mesh vertices when a body-model reconstruction is available. Let predicted and ground-truth joints be
\(\hat{\mathbf{J}}_t=\{\hat{\mathbf{j}}^{(i)}_t\}_{i=0}^{N-1}\) and
\(\mathbf{J}_t=\{\mathbf{j}^{(i)}_t\}_{i=0}^{N-1}\), with
\(\hat{\mathbf{j}}^{(i)}_t,\mathbf{j}^{(i)}_t \in \mathbb{R}^3\). We define
\begin{equation}
\mathrm{MPJPE}
=
1000 \cdot \frac{1}{N}
\sum_{i=0}^{N-1}
\left\|
\hat{\mathbf{j}}^{(i)}_t - \mathbf{j}^{(i)}_t
\right\|_2,
\end{equation}
reported in millimeters. P-MPJPE first aligns the predicted joints to the ground truth using a Procrustes similarity transform~\cite{umeyama2002least}. MPVE and P-MPVE are defined analogously on mesh vertices.

These geometry metrics are \emph{secondary} in our setting because they depend not only on pose quality but also on how body shape is handled. MPJAE is therefore the main metric for comparing IK performance, while joint and mesh errors are used to assess downstream compatibility with SMPL-style reconstruction pipelines.

\subsection{SMPL animation benchmark}
\label{sec:results_amass}

\noindent
\textbf{Protocol.}
We use AMASS~\cite{mahmood2019amass} as the primary benchmark. For each frame we recover 22 SMPL body joints in root space and compute the corresponding bone-aligned world rotations using Eq.~\ref{eq:bone_world}, yielding paired supervision targets
\((\tilde{\mathbf{J}}_t,\{\mathbf{R}^{(i)}_{t,\mathrm{bone}}\}_{i=0}^{N-1})\) with maximum canonical axis-angle invertibility error $7.96\mathrm{e}{-6}$; (mean $1.84\mathrm{e}{-6}$).
We train on the AMASS $\texttt{CMU}$ subset, validate on $\texttt{HDM05}$, $\texttt{HumanEva}$, $\texttt{MoSh}$, $\texttt{SFU}$ subsets, and report final numbers on $\texttt{SSM}$ and $\texttt{Transitions}$ subsets. The dataset comprises ~1M train frames, 0.33M validation frames, and 0.03M test frames, respectively. The model predicts pose only. Unless noted otherwise, body-model metrics for feed-forward methods are computed using ground-truth shape so that the benchmark isolates pose recovery.

\noindent
\textbf{Baselines.}
We establish a lower bound by evaluating metrics with all parameters fixed to zero, and compare against three optimization baselines (LBFGS, VPoser~\cite{ghorbani2019vposer}, and SMPLify3D~\cite{joints2smpl}) and five feed-forward baselines (MLP, QuaterGCN~\cite{song2024quatergcn}, QuaterGCN-6D, STIK, and a dense Transformer). LBFGS is identical to VPoser without the human body prior. All feed-forward models consume the same 22 root-space joints and are trained to predict the same rotation targets. Further details about the neural baselines are provided in Appendix~\ref{app:baseline-architectures}. We omit numerical comparisons with HybrIK~\cite{li2021hybrik}, MANIKIN~\cite{jiang2024manikin}, and HierarIK~\cite{yi2021hierarik} as they rely on visual cues or 6-DoF orientation priors absent in our pure 3D positional setting, as well as the lack of reproducible code for the latter two for fair evaluation.

\noindent
\textbf{Implementation details.} All models are trained on the same data splits with the same preprocessing and early-stopping criterion, so performance differences can be attributed to the model design rather than mismatched training procedures. Our default configuration uses hidden dimension \(F=256\), \(D=4\) graph-attention blocks, and 8 attention heads. We train with AdamW, batch size 512, initial learning rate 0.001, weight decay 0.01 (bias, norm and embedding has no decay), dropout 0.1, and FK-consistency weight \(\alpha=0.1\). Training runs for up to 10 epochs with early stopping if validation MPJAE does not improve for 3 consecutive epochs. Parameter counts and multiply-adds are reported for batch size 1.

\begin{figure}[b]
    \centering
    \includegraphics[width=0.9\linewidth]{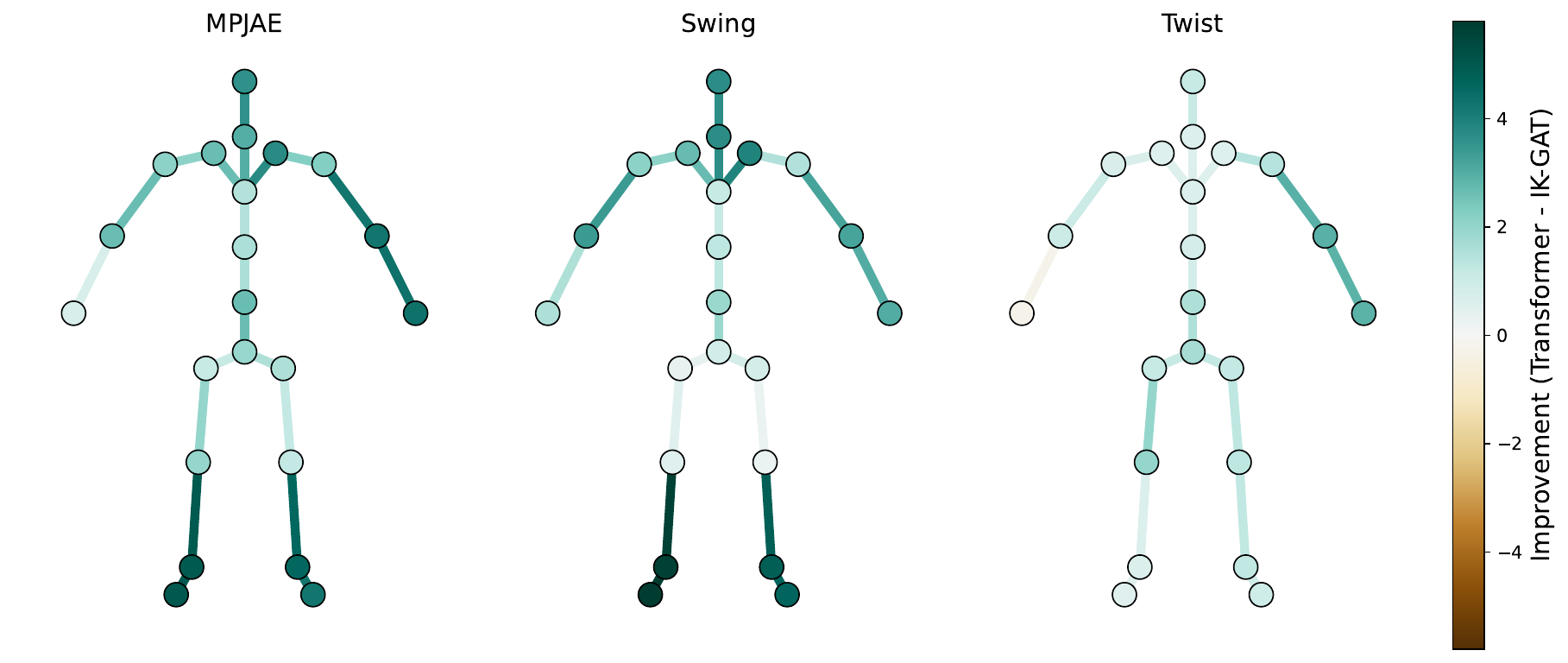}
    \caption{\textbf{Per-joint improvement over the strongest dense baseline.} Gains concentrate on distal and twist-sensitive joints such as ankles, feet, elbows, neck, and head, showing where kinematic message passing and distal local refinement matter most. Forearms remain hard because they have minimal positional evidence for twist.}
    \label{fig:per_joint_improvement}
\end{figure}

\subsubsection{Main results}
Tab.~\ref{tab:amass_results} shows that the main advantage of \model\ is not only lower error, but a better alignment between the learned representation and the actual IK objective. The strongest generic baseline is the dense Transformer, which confirms that full-body rotation recovery requires global context and cannot be solved by independent joint-wise regression. However, once global interaction is already available, unrestricted attention is still suboptimal: \model\ improves test MPJAE from 9.06 to 7.43 and test MPVE from 33.08 to 24.47 while using only 0.37M parameters instead of 3.17M. This indicates that the remaining gap is not due to insufficient model capacity, but due to the absence of the two priors encoded by \model: a canonical bone-aligned rotation space and anatomically constrained message passing.
Among the iterative methods, VPoser is strongest on MPJAE, while SMPLify3D is stronger on geometry. This split reflects an objective mismatch: optimization can exploit pose-shape trade-offs to improve joint and mesh reconstruction without necessarily recovering better local rotations. Since our goal is animation-ready IK rather than generic body-model fitting, MPJAE is the more informative metric here. Under that metric, \model\ outperforms all iterative baselines despite not performing any per-frame optimization. In other words, amortization is not only reducing runtime; it is recovering a better pose variable for the task we actually care about.
A third pattern is the unusually large gap between weak and strong neural baselines. MLP fails catastrophically, and even QuaterGCN-6D remains far behind the Transformer and \model. This suggests that the hard part of dense IK is not simply rotation parameterization, but structured disambiguation: the network must decide which joints should influence which others, and in what coordinate system the prediction problem is easiest to solve. \model\ performs best because it resolves both issues simultaneously. The graph restricts information flow to plausible anatomical pathways, while bone-world supervision removes rig-specific basis ambiguity from the target space.

\noindent 
\textbf{Per-joint comparisons.}
Fig~\ref{fig:per_joint_improvement} compares \model\ with the strongest non-graph baseline at the joint level. The largest improvements appear on distal and twist-sensitive joints, where local evidence is weakest and ambiguity must be resolved from broader kinematic context. Gains are smaller on pelvis and hips because their orientations are already strongly constrained by adjacent joints. Wrists remain difficult for all methods as expected: leaf nodes provide almost no positional evidence for twist. This distribution mirrors the design of \model\ itself. The bidirectional graph propagates context along the chain, while the distal local refinement explicitly targets the joints where independent regression fails most.

\begin{figure}[htbp]
\centering
\scriptsize

\pgfplotstableread[col sep=&, row sep=\\]{
iter & mpjae & mpjpe & p_mpjpe & mpve & p_mpve \\
1    & 21.14 & 419.69 & 120.99 & 528.52 & 144.10 \\
10   & 19.02 &  92.34 &  71.59 & 146.15 &  91.80 \\
50   & 14.70 &  43.17 &  19.49 &  58.32 &  35.29 \\
100  & 14.47 &  41.12 &  12.99 &  53.94 &  32.54 \\
200  & 14.93 &  40.42 &  11.30 &  51.75 &  30.79 \\
300  & 15.14 &  39.78 &  11.20 &  51.20 &  30.49 \\
}\lbfgstable

\pgfplotstableread[col sep=&, row sep=\\]{
iter & mpjae & mpjpe & p_mpjpe & mpve & p_mpve \\
1    & 20.15 & 428.07 & 118.90 & 541.35 & 142.47 \\
10   & 16.97 & 124.07 & 105.56 & 162.08 & 120.33 \\
50   & 11.40 &  58.48 &  42.64 &  84.60 &  56.18 \\
100  &  9.98 &  47.71 &  29.55 &  69.00 &  44.91 \\
200  &  9.24 &  44.13 &  23.07 &  59.57 &  38.35 \\
300  &  9.54 &  42.58 &  21.07 &  56.71 &  36.07 \\
}\lbfgsvposerstable

\pgfplotstableread[col sep=&, row sep=\\]{
iter & mpjae & mpjpe & p_mpjpe & mpve & p_mpve \\
1    & 20.09 & 289.34 & 117.95 & 373.78 & 141.28 \\
10   & 19.13 & 133.12 & 103.13 & 184.40 & 126.77 \\
50   & 15.50 &  58.44 &  34.80 &  79.43 &  48.94 \\
100  & 12.84 &  41.95 &  20.53 &  57.63 &  37.33 \\
200  & 11.86 &  34.61 &  14.03 &  48.91 &  31.99 \\
300  & 12.29 &  31.28 &  12.66 &  44.72 &  30.05 \\
}\smplifythreedtable

\begin{tikzpicture}
\begin{groupplot}[
  group style={
    group size=3 by 1,
    horizontal sep=0.12\linewidth
  },
  width=0.22\linewidth,
  height=1.7cm,
  xmode=log,
  xmin=1, xmax=300,
  scale only axis,
  tick label style={font=\scriptsize},
  xticklabel style={font=\scriptsize},
  yticklabel style={font=\scriptsize},
  label style={font=\scriptsize},
  legend columns=4,
  legend style={
    font=\scriptsize,
    draw=none,
    at={(2.0,1.02)},
    anchor=south,
    /tikz/every even column/.append style={column sep=0.2cm}
  },
]

\nextgroupplot[
  ylabel=MPJAE,
  xlabel=Num. Iterations,
  xticklabels={1,10,100},
]
\addplot[black, solid, thick] coordinates {(1,7.43) (300,7.43)};
\addlegendentry{IK-GAT}
\addplot[blue, mark=o] table[x=iter, y=mpjae] {\lbfgstable};
\addlegendentry{LBFGS}
\addplot[red, mark=square] table[x=iter, y=mpjae] {\lbfgsvposerstable};
\addlegendentry{VPoser}
\addplot[green!60!black, mark=triangle] table[x=iter, y=mpjae] {\smplifythreedtable};
\addlegendentry{SMPLify3D}
\addplot[black, dashed, thick] coordinates {(200,7) (200,21)};

\nextgroupplot[
  ylabel=MPJPE,
  xlabel=Num. Iterations,
  xticklabels={1,10,100},
]
\addplot[black, dashed, thick] coordinates {(200,31) (200,428)};
\addplot[black, solid, thick] coordinates {(1,47.45) (300,47.45)};
\addplot[blue, mark=o] table[x=iter, y=mpjpe] {\lbfgstable};
\addplot[red, mark=square] table[x=iter, y=mpjpe] {\lbfgsvposerstable};
\addplot[green!60!black, mark=triangle] table[x=iter, y=mpjpe] {\smplifythreedtable};

\nextgroupplot[
  ylabel=MPVE,
  xlabel=Num. Iterations,
  xticklabels={1,10,100},
]
\addplot[black, dashed, thick] coordinates {(200,44) (200,542)};
\addplot[black, solid, thick] coordinates {(1,63.34) (300,63.34)};
\addplot[blue, mark=o] table[x=iter, y=mpve] {\lbfgstable};
\addplot[red, mark=square] table[x=iter, y=mpve] {\lbfgsvposerstable};
\addplot[green!60!black, mark=triangle] table[x=iter, y=mpve] {\smplifythreedtable};

\end{groupplot}
\end{tikzpicture}

\caption{\textbf{Iteration budget versus accuracy on AMASS.} Optimization baselines need tens to hundreds of iterations to approach the accuracy reached by a single forward pass of IK-GAT.}
\label{fig:ik_tradeoff}
\end{figure}
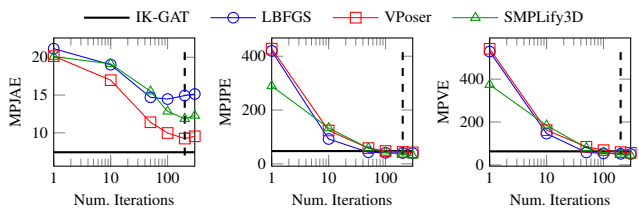

\noindent 
\textbf{Iteration-budget tradeoff.}
Fig~\ref{fig:ik_tradeoff} isolates the cost of iterative fitting. The optimization curves only approach \model\ after tens to hundreds of iterations, and their gains beyond that point are larger on joint and mesh metrics than on local-space MPJAE. This divergence is informative: once shape and global body configuration can absorb residual error, geometry can continue improving even when local rotations do not. In contrast, \model\ directly targets the animation variable of interest.

\noindent
\textbf{Runtime.} Single-frame inference reaches 693 FPS on CPU and 448 FPS on GPU (RTX 4060), while GPU provides up to \(3.7\times\) higher throughput under batching (Appendix~\ref{app:runtime_analysis}).



\subsubsection{Ablation study}
\label{sec:ablations}


\noindent
\textbf{Effect of bone-aligned world rotation space.} Training directly in parent-relative local space forces the network to learn both pose and rig-specific basis conventions at the same time (Tab~\ref{tab:ablation_rotation_space}). Bone-aligned world space removes this nuisance variable by canonicalizing the local frame once, outside the network. The resulting gains are not limited to MPJAE. They are even larger on MPJPE and MPVE after local recovery, which indicates that the representation improves global kinematic consistency rather than merely making angular regression numerically easier.

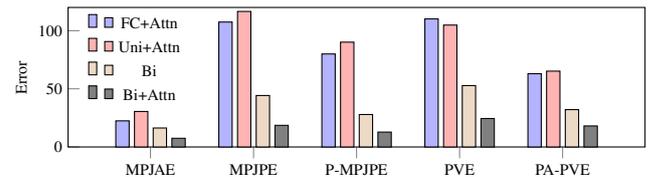
\begin{figure}[htbp]
\centering
\scriptsize


\pgfplotstableread[row sep=\\,col sep=&]{
Metric     & FC    & Uni    & Bi    & BiAttn \\
MPJAE      & 22.45 & 30.53  & 16.31 & 7.48   \\
MPJPE      &107.45 & 116.57 & 44.26 & 18.61  \\
P-MPJPE   & 80.00 & 90.16  & 27.92 & 12.80  \\
PVE        &110.14 & 104.84 & 52.80 & 24.48  \\
PA-PVE     & 63.00 & 65.23  & 32.11 & 18.08  \\
}\graphdata

\begin{tikzpicture}c
\begin{axis}[
    scale only axis,
    width=0.9\linewidth,
    height=1.85cm,
    ybar,
    bar width=5pt,
    ymin=0,
    ymax=120,
    ylabel={Error},
    symbolic x coords={MPJAE,MPJPE,P-MPJPE,PVE,PA-PVE},
    xtick=data,
    legend style={
        at={(0.02,0.625)},
        anchor=west,
        legend columns=1,
        draw=none,
        fill=none,
    },
    xtick pos=bottom,
    ytick pos=left,
    enlarge x limits=0.2
]

\addplot+[draw=black] table[x=Metric, y=FC] {\graphdata};
\addlegendentry{FC+Attn}

\addplot+[draw=black] table[x=Metric, y=Uni] {\graphdata};
\addlegendentry{Uni+Attn}

\addplot+[draw=black] table[x=Metric, y=Bi] {\graphdata};
\addlegendentry{Bi}

\addplot+[draw=black] table[x=Metric, y=BiAttn] {\graphdata};
\addlegendentry{Bi+Attn}

\end{axis}
\end{tikzpicture}

\caption{\textbf{Ablation of graph structure and attention.} Bidirectional kinematic message passing with attention is critical. Fully connected graphs dilute the anatomical inductive bias, one-way graphs block corrective feedback along the chain, and non-attentive graph propagation cannot adapt neighbor importance to the current pose.}
\label{fig:ablation-graph}
\end{figure}

\noindent \textbf{Effect of graph structure and attention.}
Fig~\ref{fig:ablation-graph} shows that graph structure is a first-order design choice. Fully connected attention weakens the anatomical bias by encouraging interactions between unrelated joints, while a unidirectional graph prevents information from propagating back along the chain. Replacing attention with standard graph convolution also hurts performance, as neighbors are then treated too uniformly. Best result comes from the combination of bidirectional kinematic graph and learned attention weights.

\begin{table}[htbp]
\small
\setlength{\tabcolsep}{3pt}
\caption{\textbf{Ablation of rotation space.} Training in bone-aligned world space outperforms direct local-rotation regression, especially on geometry-aware metrics, showing that the proposed representation simplifies learning without sacrificing accuracy.}
\label{tab:ablation_rotation_space}
\begin{tabularx}{\linewidth}{|X|ccccc|}
\hline
\textbf{Rotation Space} & \textbf{MPJAE} & \textbf{MPJPE} & \textbf{P-MPJPE} & \textbf{MPVE} & \textbf{P-MPVE} \\
\hline
Parent-Relative & 7.93 & 31.68 & 23.49 & 47.31 & 29.29 \\
\rowcolor{gray!20}
Bone-World      & 7.48 & 18.61 & 12.80 & 24.48 & 18.08 \\
\hline
\end{tabularx}
\end{table}

\noindent \textbf{Effect of \model~components.}
Tab~\ref{tab:ablation-components} shows that positional embeddings are indispensable: without joint identity, the network struggles to distinguish anatomically distinct but geometrically similar configurations, and MPJAE degrades sharply. Once embeddings are present, both shortcut mechanisms offer smaller but complementary gains. The global shortcut preserves direct access to the observed geometry, while the local refinement improves the parts of the body where local ambiguity accumulates most strongly.

\begin{table}[htbp]
\centering
\small
\setlength{\tabcolsep}{3pt}
\caption{\textbf{Component ablation under the bidirectional graph.} Positional embeddings (\textbf{PE}) are essential as they disambiguate anatomically distinct joints under similar local geometry. The global shortcut (\textbf{GS}) and distal local refinement (\textbf{LR}) provide smaller but complementary gains, with the full model giving the best balance.}
\label{tab:ablation-components}
\begin{tabularx}{\linewidth}{|YYY|ccccc|}
\hline
\textbf{PE} & \textbf{GS} & \textbf{LR} & \textbf{MPJAE} & \textbf{MPJPE} & \textbf{P-MPJPE} & \textbf{MPVE} & \textbf{P-MPVE} \\
\hline
\checkmark &            &            &  7.58 & 19.02 & 13.65 & \textbf{24.29} & \underline{18.40} \\
           & \checkmark &            & 15.68 & 44.94 & 33.09 & 57.83 & 38.34 \\
           &            & \checkmark & 16.31 & 45.09 & 34.24 & 56.80 & 38.24 \\
           & \checkmark & \checkmark & 17.26 & 45.00 & 33.13 & 54.78 & 37.08 \\
\checkmark &            & \checkmark &  \underline{7.51} & \textbf{18.01} & 13.58 & 24.72 & 19.31 \\
\checkmark & \checkmark &            &  7.64 & 19.04 & \textbf{12.75} & 25.87 & 18.65 \\
\rowcolor{gray!20}
\checkmark & \checkmark & \checkmark &  \textbf{7.48} & \underline{18.61} & \underline{12.80} & \underline{24.48} & \textbf{18.08} \\
\hline
\end{tabularx}
\end{table}

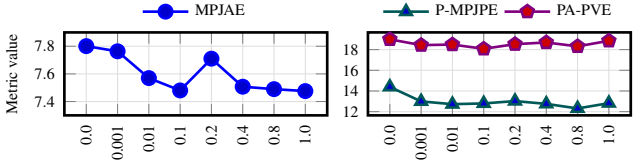
\begin{figure}[htbp]
  \centering

  \pgfplotstableread{

Alpha           MPJAE   PA_MPJPE        PA_PVE
0.0             7.8008  14.3950         18.9957
0.001           7.7633  12.9920         18.4336
0.01            7.5696  12.7239         18.4889
0.1             7.4803  12.7960         18.0774
0.2             7.7101  13.0235         18.5260
0.4             7.5073  12.7533         18.6914
0.8             7.4897  12.3110         18.3142
1.0             7.4756  12.8220         18.8533
  }\alphadata

  \begin{tikzpicture}
  \begin{axis}[
      scale only axis,
      width=0.41\linewidth,
      height=1.1cm,
      ylabel={Metric value},
      symbolic x coords={0.0,0.001,0.01,0.1,0.2,0.4,0.8,1.0},
      xtick=data,
      ymin=7.3,
      ymax=7.9,
      grid=both,
      major grid style={gray!25},
      minor grid style={gray!15},
      tick label style={font=\scriptsize},
      x tick label style={rotate=90,anchor=east},
      label style={font=\scriptsize},
      legend style={
          at={(0.5,1.02)},
          anchor=south,
          legend columns=3,
          draw=none,
          font=\scriptsize
      },
      line width=1pt,
      mark size=2.5pt,
  ]

  \addplot+[color=blue, mark=*]
  table[x=Alpha,y=MPJAE] {\alphadata};
  \addlegendentry{MPJAE}

  \end{axis}
  \end{tikzpicture}
  \begin{tikzpicture}
  \begin{axis}[
      scale only axis,
      width=0.41\linewidth,
      height=1.1cm,
      symbolic x coords={0.0,0.001,0.01,0.1,0.2,0.4,0.8,1.0},
      xtick=data,
      grid=both,
      major grid style={gray!25},
      minor grid style={gray!15},
      tick label style={font=\scriptsize},
      label style={font=\small},
      x tick label style={rotate=90,anchor=east},
      legend style={
          at={(0.5,1.02)},
          anchor=south,
          legend columns=3,
          draw=none,
          font=\scriptsize
      },
      line width=1pt,
      mark size=2.5pt,
  ]

  \addplot+[color=teal!70!black, mark=triangle*]
  table[x=Alpha,y=PA_MPJPE] {\alphadata};
  \addlegendentry{P-MPJPE}

  \addplot+[color=violet, mark=pentagon*]
  table[x=Alpha,y=PA_PVE] {\alphadata};
  \addlegendentry{PA-PVE}

  \end{axis}
  \end{tikzpicture}
  \caption{\textbf{Effect of FK-consistency weight \(\alpha\).} A small FK regularization improves both local-rotation accuracy and downstream geometry by tying the predicted rotations back to the observed joints. Larger weights over-constrain training and make the optimization trade-off increasingly unstable, so we use \(\alpha=0.1\) throughout.}
  \label{fig:weight-search}
\end{figure}

\noindent \textbf{Effect of FK-consistency loss.}
Fig~\ref{fig:weight-search} studies the influence of the FK-consistency weight \(\alpha\). A small auxiliary term is beneficial because it couples the predicted rotations back to the observed joint geometry and discourages locally plausible but globally inconsistent solutions. However, once \(\alpha\) becomes too large, training starts to chase joint reconstruction too aggressively and the trade-off between rotational and geometric metrics becomes unstable. We therefore use \(\alpha=0.1\) as the most reliable compromise.

\begin{figure}[htbp]
    \centering
    \scriptsize
    \begin{tikzpicture}
    \pgfplotstableread{
    Name   MPJAE  MPJPE  PVE    Params
    Tiny   22.13  49.91  50.64   17.2
    Small  10.11  21.15  25.26   80.1
    Base    7.48  12.80  18.08  374.0
    Large   7.61  12.54  19.41  981.0
    Huge    7.72  12.46  20.78 2000.0
    }\modeldata
    
    \begin{axis}[
        width=\linewidth,
        height=3.0cm,
        xmode=log,
        xlabel={Parameters (K)},
        ylabel={Error},
        grid=major,
        legend style={
            draw=none,
            fill=none,
            at={(0.98,0.99)},
            legend columns=-1,
            anchor=north east
        },
    ]
    
    \addplot+[mark=*, thick] coordinates {
        (17.2,22.13)
        (80.1,10.11)
        (374.0,7.48)
        (981.0,7.61)
        (2000.0,7.72)
    };
    \addlegendentry{MPJAE}
    
    \addplot+[mark=square*, dashed, thick] coordinates {
        (17.2,49.91)
        (80.1,21.15)
        (374.0,12.80)
        (981.0,12.54)
        (2000.0,12.46)
    };
    \addlegendentry{P-MPJPE}
    
    \addplot+[mark=triangle*, dashdotted, thick] coordinates {
        (17.2,50.64)
        (80.1,25.26)
        (374.0,18.08)
        (981.0,19.41)
        (2000.0,20.78)
    };
    \addlegendentry{PA-PVE}
    
    \end{axis}
    \end{tikzpicture}
    \caption{\textbf{Model scaling.} Performance improves rapidly from Tiny to Base, then largely saturates. This indicates that the main gains come from the representation and the anatomical inductive bias rather than from scaling capacity alone.}
    \label{fig:size-tradeoff}
\end{figure}
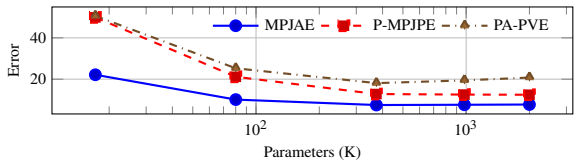

\noindent \textbf{Effect of model size.}
Fig~\ref{fig:size-tradeoff} shows a clear diminishing-return regime. Scaling from Tiny to Base (default) yields the expected improvement, but larger models do not improve the primary MPJAE metric and even slightly worsen MPVE. Once the representation and the kinematic prior are fixed, additional capacity does not resolve the underlying ambiguity more effectively.


\subsection{Game-rig animation benchmark}
\label{sec:results_ue}

The second benchmark addresses a different question: not whether the recovered pose best explains a parametric body model, but whether the analytically recovered local rotations are directly usable on a production rig in a realistic deployment setting. We therefore evaluate \model\ on the UE5 mannequin, with no body model in the loop and no retargeting at inference time.

\begin{figure}[t]
\centering
\includegraphics[width=\linewidth]{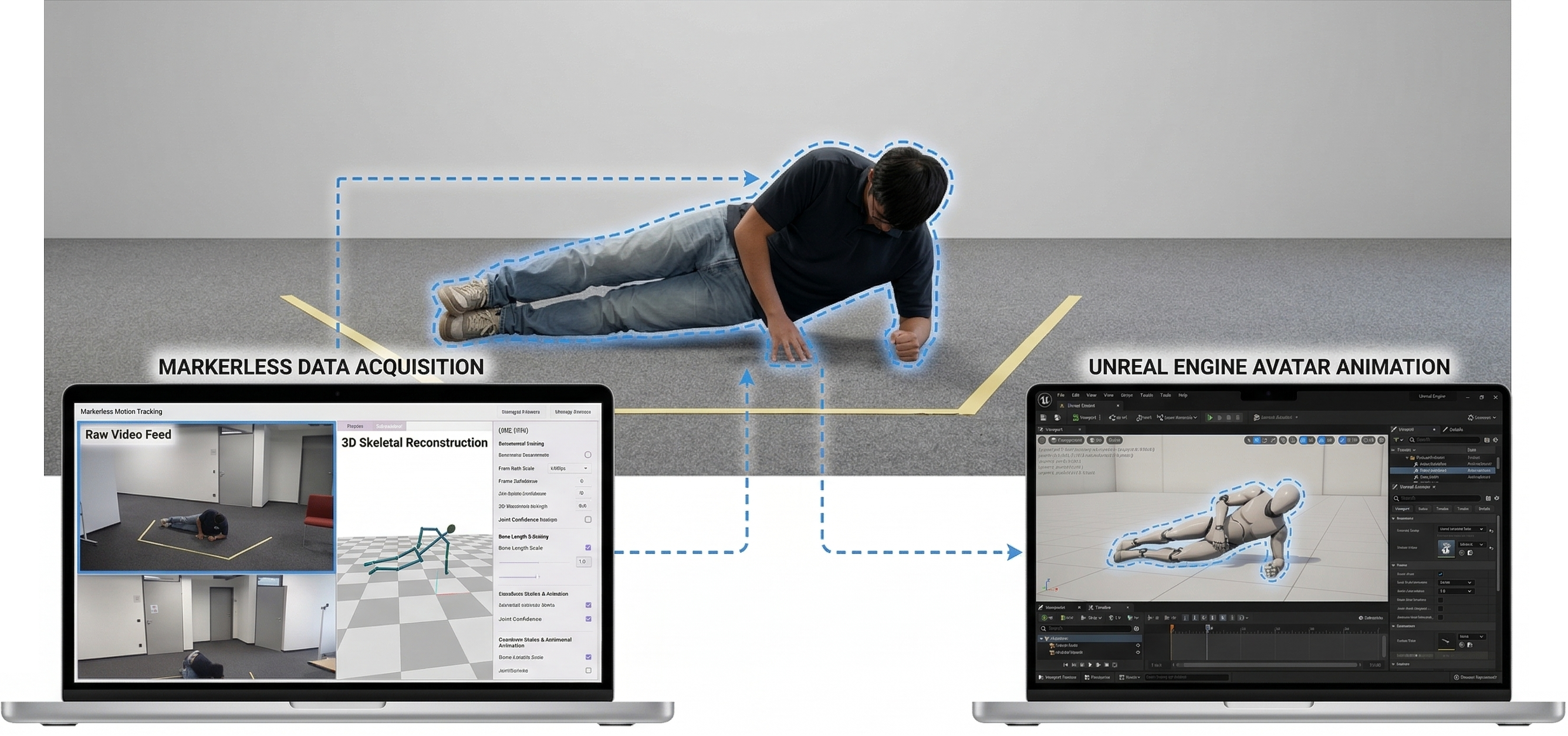}
\caption{\textbf{Real-time deployment setting.} A tracked human motion is first converted into a dense stream of 3D joints by an upstream markerless multi-view system, then mapped by \model\ to per-joint rotations in bone-aligned world space, analytically recovered to UE5-Mannequin local rotations, and rendered directly in Unreal Engine. This benchmark evaluates the complete animation-facing inverse-kinematics pipeline rather than body-model reconstruction.}
\label{fig:ue_deployment}
\end{figure}

\noindent
\textbf{Benchmark setup and motivation.}
Our target application is remote physiotherapy~\cite{coraggio2023embodied}, where a tracked person performs full-body exercises and the motion is streamed to an avatar in Unreal Engine. This setting is challenging for two reasons. First, many target motions, especially low-posture rehabilitation exercises such as planks and bridges, are underrepresented in public motion datasets. Second, public Unreal-oriented datasets~\cite{akada2022unrealego} often target older skeleton conventions, so retargeting would introduce an additional source of error that is orthogonal to the inverse-kinematics problem studied here. We therefore construct a benchmark directly in the UE5 mannequin convention and evaluate whether \model\ transfers beyond the standardized SMPL setting. In end-to-end deployment illustrated in Fig~\ref{fig:ue_deployment}, a multi-view markerless tracker~\cite{khan2026simspine} provides dense 3D joints, \model\ converts these joints into animation-ready local rotations, and the recovered rotations are streamed to Unreal Engine for real-time rendering.

\begin{table}[htbp]
\small
\setlength{\tabcolsep}{3pt}
\caption{\textbf{UE-IK dataset statistics.} The training set combines clean marketplace animations with real-world IMU-driven UE5 sequences, while validation and test emphasize rig-specific articulation and deployment-relevant motion. This split lets us test whether a model trained largely on clean synthetic animation can transfer to real recorded exercises in the same target skeleton convention.}
\label{tab:ue_dataset_stats}
\begin{tabularx}{\linewidth}{|l|p{1.2cm}X|}
    \hline
    \textbf{Split} & \textbf{Sequences} / \textbf{Frames} & \textbf{Notes} \\
    \hline
    Train & 491 / 305k & 80\% of our IMU-based real-world data (59 sequences, 158k frames), combined with marketplace animations covering dancing, gym workouts, yoga, locomotion, walking, and other everyday motions. \\
    Val   & 40 / 1k    & Built-in UE5 male (Manny) and female (Quinn) avatars, each articulating individual bones through their full range of motion. \\
    Test  & 17 / 49k   & 20\% of the real-world IMU-based data, including free movements, target rehabilitation workouts, and range-of-motion sequences. \\
    \hline
\end{tabularx}
\end{table}

\begin{figure}[t]
\centering
\includegraphics[width=\linewidth]{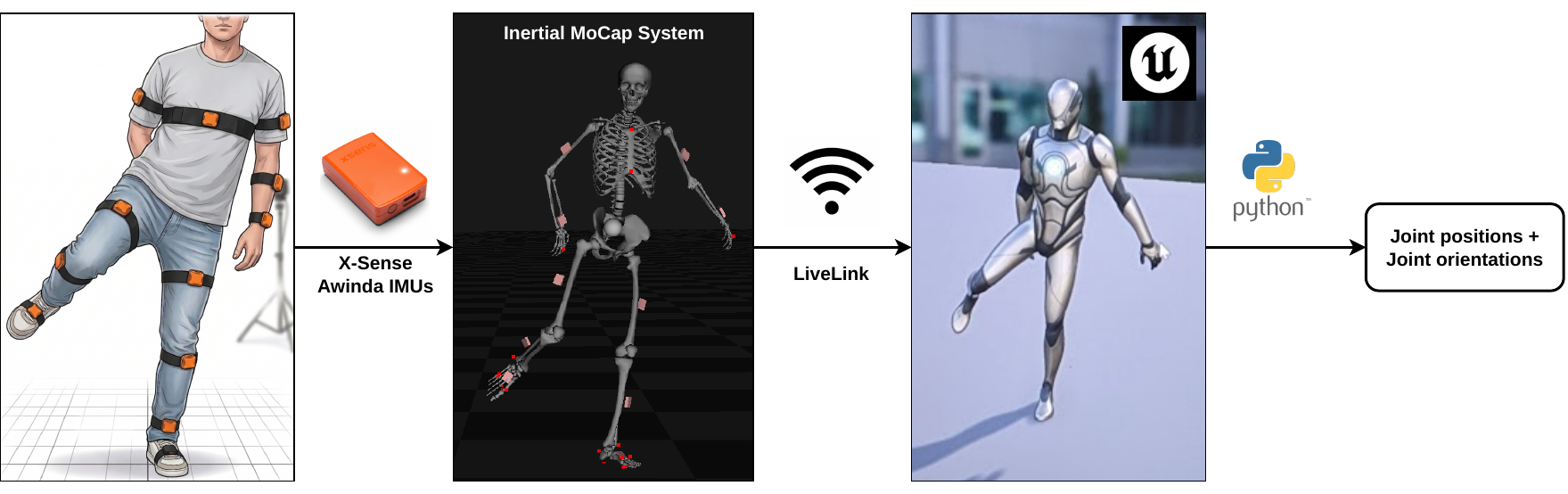}
\caption{\textbf{UE-IK acquisition process.} Real motion is captured with Xsens IMUs, streamed to UE, and exported as paired joint positions and rotations. Marketplace animations are exported through the same pipeline, giving unified training data for the benchmark.}
\label{fig:ue_capture_pipeline}
\end{figure}

\noindent
\textbf{UE-IK dataset construction.}
To train and evaluate the method in this rig convention, we construct the \emph{UE-IK} dataset comprising paired root-space joint positions and local joint rotations in the UE5 mannequin skeleton. The data, summarized in Tab~\ref{tab:ue_dataset_stats}, combines two complementary sources. First, we collect a large set of high-quality animation clips from Unreal Engine's marketplace and export per-frame joint positions and local bone rotations through the Unreal Python API. These clips provide broad motion diversity and clean supervision in the exact target skeleton convention. Second, to better reflect the target rehabilitation scenario, we record real people using an Xsens IMU-based motion-capture system~\cite{taetz2025jointtracker}, stream the motion to the UE5 mannequin through LiveLink, and export the resulting supervision data in the same format. This second source introduces motions that are both closer to deployment and less well represented in generic public datasets.
The real-world recording protocol, illustrated in Fig~\ref{fig:ue_capture_pipeline}, covers twelve subjects. Each subject performs three types of motion: full-range-of-motion articulation for major body parts, repeated target rehabilitation exercises, and short free-form movement sequences.

\noindent
\textbf{Training protocol and transfer.}
All models are initialized from their corresponding AMASS-pretrained checkpoints and then adapted to the UE-IK benchmark. This transfer step is important because UE-IK is much smaller than the SMPL benchmark and differs not only in motion distribution but also in skeleton topology and joint count. To make transfer effective across skeletons, we remap learned positional embeddings from the source skeleton to the corresponding destination joints, initialize unmatched joints with zeros, and partially reuse pretrained weights for layers whose structure depends on the number of joints. This preserves a large fraction of the learned anatomical prior and substantially improves convergence under limited target data.

\begin{table}[t]
\centering
\small
\setlength{\tabcolsep}{3pt}
\caption{\textbf{UE5 mannequin benchmark.} All models are initialized from their corresponding AMASS-pretrained checkpoints and fine-tuned on UE-IK. Lower is better. IK-GAT achieves the best overall local-space MPJAE as well as the lowest swing and twist errors, showing that the proposed representation and kinematic message passing remain effective after transfer to a production rig.}
\label{tab:results-ue-ik-val}
\begin{tabularx}{\linewidth}{|l|YYY|Y|}
\hline
\textbf{Method} & \textbf{MPJAE} & \textbf{Swing} & \textbf{Twist} & \textbf{Params(M)} \\
\hline
MLP             & 75.08 & 38.27 & 57.79 & 2.11 \\
QuaterGCN-6D    & 30.33 & 22.01 & 19.29 & 0.13 \\
STIK            & 20.64 & 12.15 & 14.25 & 1.32 \\
Transformer     & 15.61 &  7.39 & 11.93 & 3.17 \\
\rowcolor{gray!20}
\model~(ours)   & \textbf{12.83} &  \textbf{5.20} & \textbf{10.42} & 0.37 \\
\hline
\end{tabularx}
\end{table}

\begin{figure}[t]
\centering
\includegraphics[width=\linewidth]{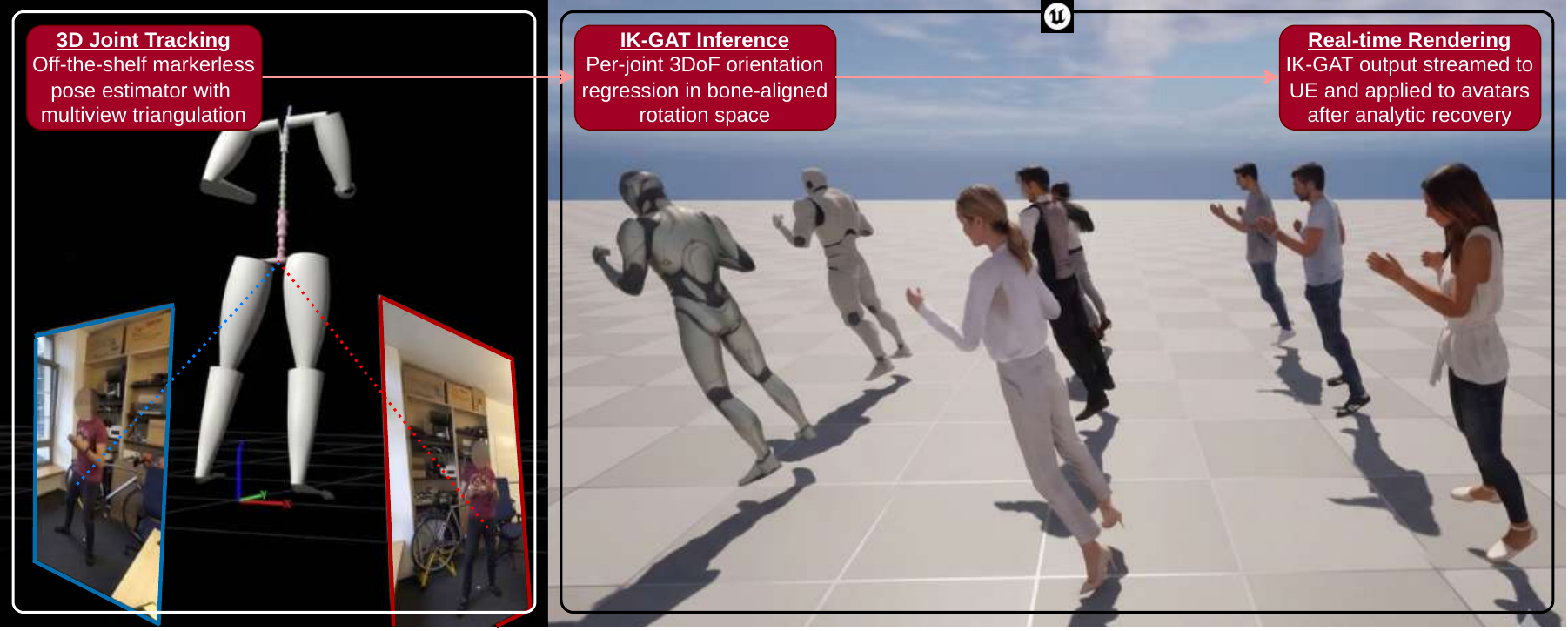}
\caption{\textbf{UE5 qualitative deployment.} Example frame rendered directly from \model\ predictions on multiple UE5 avatars.}
\label{fig:ue_qualitative}
\end{figure}

\noindent
\textbf{Main results.}
Tab~\ref{tab:results-ue-ik-val} shows that the ranking on the UE5 benchmark mirrors the main SMPL study, but with a sharper separation between models that explicitly encode structure and those that do not. The MLP baseline fails by a wide margin, which reinforces the central claim of the paper: dense position-to-dense orientation recovery is not a simple regression task. Even when every joint is observed, the output must satisfy a coupled kinematic structure and resolve twist ambiguity in a rig-consistent way. Purely token-wise regression is therefore insufficient. Introducing more structure improves performance progressively. QuaterGCN-6D is substantially stronger than the MLP, indicating that graph-based propagation is already beneficial. STIK improves further by adding temporal and parent-conditioned modeling, which is consistent with the intuition that even weak kinematic cues are useful when data is limited. The Transformer is the strongest baseline, showing that global context is important on this benchmark. However, \model\ still achieves the best result despite using a much smaller parameter budget than the Transformer. This is the key result of the UE benchmark: once the target is a real production rig, anatomically constrained message passing and exact local-rotation recovery are more valuable than generic dense attention alone. The gap is also informative when decomposed into swing and twist. The improvement in swing indicates that the graph prior helps reconstruct globally coherent limb directions under cross-skeleton transfer. The remaining gain in twist, although smaller in absolute terms, is the more meaningful one: twist is exactly the part of the problem that is weakly constrained by positions alone, especially under distribution shift. Fig~\ref{fig:ue_qualitative} complements the table by qualitatively demonstrating that the recovered local rotations can be used as-is inside Unreal Engine, preserving chain coherence and plausible articulation without any additional post-processing step.

\section{Conclusion}
\label{sec:conclusion}

We presented \model, a lightweight inverse-kinematics model that reconstructs full-body joint orientations from 3D joint positions in a single forward pass. The key idea is to separate the learning problem from the kinematic conversion: the network predicts bone-aligned world rotations, while the transformation to standard parent-relative joint rotations is recovered analytically from the known rest pose and skeletal hierarchy. This decomposition reduces the complexity of the learned component and yields a representation that is stable for regression while remaining exactly compatible with conventional animation pipelines.
Across both standardized and deployment-oriented benchmarks, the results support the same conclusion: dense-joint IK benefits most from the combination of a canonical rotation space, an explicit anatomical prior, and exact recovery to local rotations. The gains are especially pronounced on distal and twist-sensitive joints, where generic feed-forward regression and per-sample optimization are weakest. Because the method does not require iterative fitting or body-model optimization at inference time, it is well suited for online avatar animation, telepresence, live virtual production, and interactive character control.
The current formulation still inherits three limitations of position-only IK: twist remains intrinsically ambiguous in rare or out-of-distribution poses, the analytic recovery assumes a known rest pose and skeletal topology, and the default model is single-frame rather than explicitly temporal. These limitations point to the most promising next steps: short-horizon temporal context, additional sensing cues such as IMUs or surface orientation, and topology-robust transfer across heterogeneous rigs.

\section*{Acknowledgment}

This work was co-funded by the European Union’s Horizon Europe research and innovation programme under Grant Agreement No 101135724 (LUMINOUS).



\printbibliography

\clearpage
\appendix
\renewcommand{\thefigure}{\thesection\arabic{figure}}
\renewcommand{\thetable}{\thesection\arabic{table}}

\section{Additional Results}

This appendix provides supplementary results supporting the main-paper claims, with additional evidence on UE5 transfer and robustness under corrupted inputs.

\subsection{Additional UE5 qualitative result}
\label{app:qualitative_extra}

\begin{figure}[htbp]
\centering
\includegraphics[width=\linewidth]{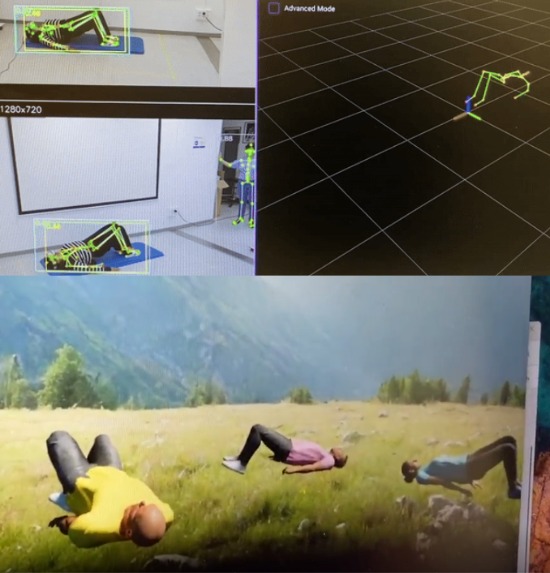}
\caption{\textbf{UE5 qualitative result on a low-posture exercise.} Top: tracked pose input and reconstructed skeleton. Bottom: UE5 avatar animation driven directly by \model\ predictions after analytic recovery to local rotations. This complements the main-paper UE5 figure by focusing on a rehabilitation-style motion that is rare in standard datasets and more sensitive to articulation errors.}
\label{fig:ue_qualitative_appendix}
\end{figure}

Fig~\ref{fig:ue_qualitative_appendix} complements the UE5 benchmark in the paper with a more challenging rehabilitation-style motion. It shows the tracked pose and reconstructed skeleton used as input to \model at top, and the resulting UE5 avatar animation driven by analytically recovered local joint rotations at bottom. Consistent with Tab~\ref{tab:results-ue-ik-val}, the transferred model preserves chain coherence and produces directly usable rig motion without retargeting or body-model fitting, even in poses where joint-to-rotation recovery is more ambiguous.

\subsection{Additional SMPL qualitative result}
\label{app:qualitative_extra_smpl}

\begin{figure}[htbp]
    \centering
    \includegraphics[height=3cm]{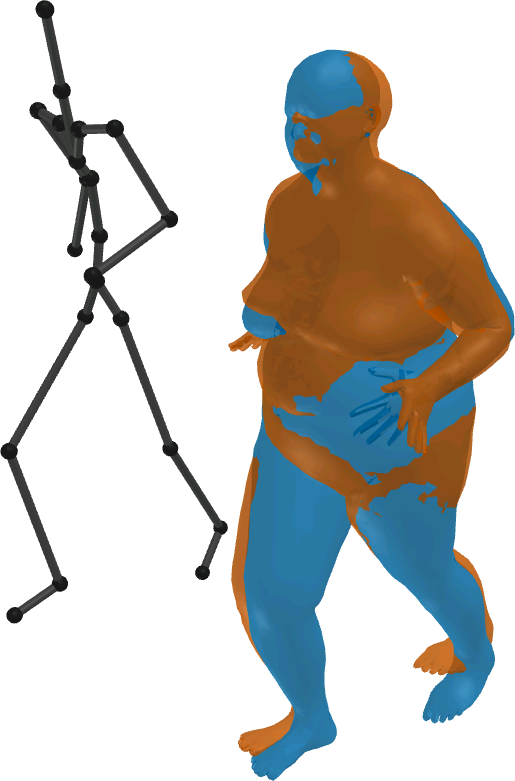}
    \includegraphics[height=3cm]{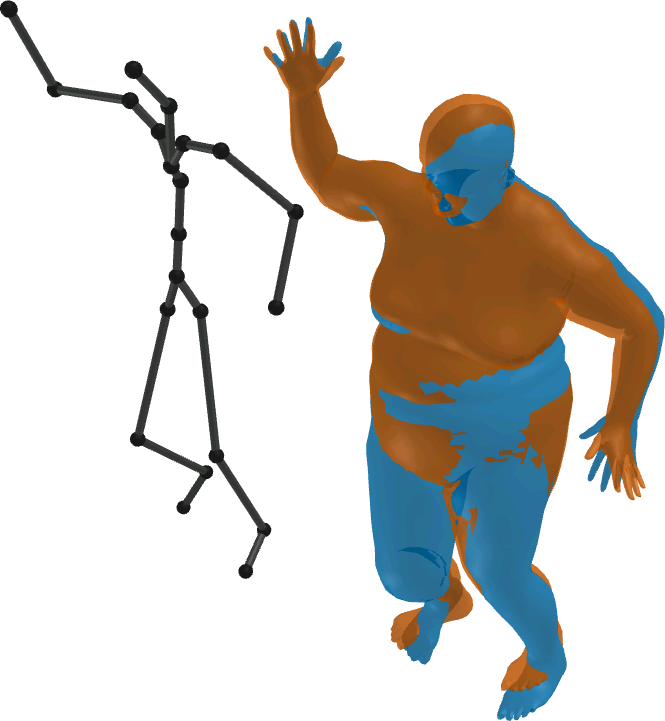}
    \includegraphics[height=3cm]{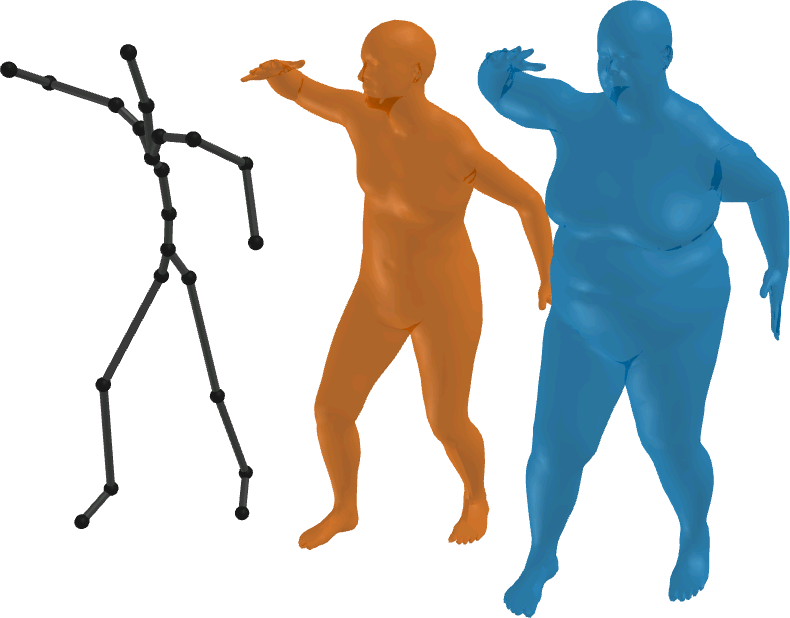}
    \includegraphics[height=3cm]{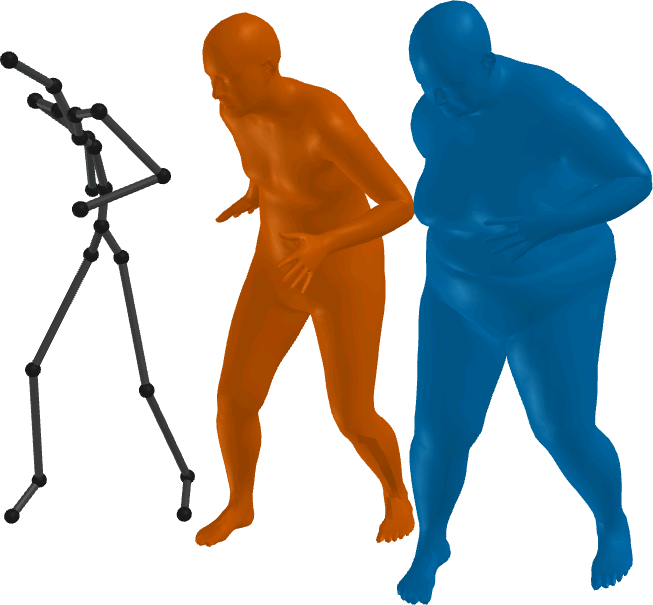}
    \includegraphics[height=2.5cm]{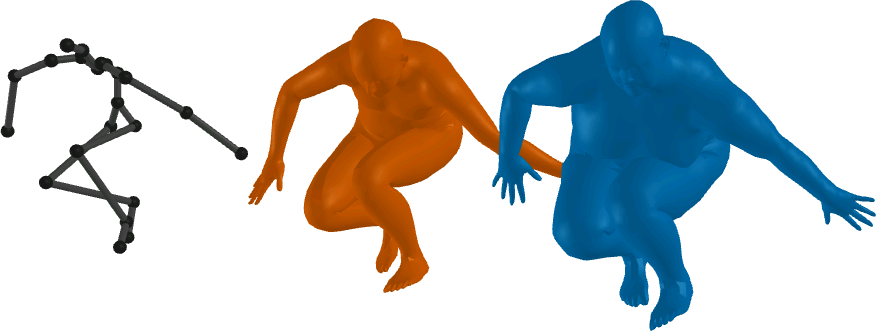}
    \caption{\textbf{SMPL qualitative results.} Input 3D joints (black) and reconstructed meshes (brown) compared to GT (blue). Top row uses GT shape, while other rows use fixed shape across samples.}
    \label{fig:smpl_qualitative_appendix}
\end{figure}

Fig~\ref{fig:smpl_qualitative_appendix} shows reconstruction results on the SMPL benchmark. Although \model\ takes only 3D joint positions as input, the recovered poses align closely with the ground-truth mesh. With ground-truth shape parameters (top row), the remaining error is mainly due to small pose inaccuracies. With fixed shape (bottom rows), visible shape mismatch appears, but articulation remains consistent. This indicates that \model\ primarily learns a stable inverse-kinematics mapping rather than relying on implicit shape cues.

\subsection*{Robustness to Gaussian positional noise}
\label{app:gaussian_noise_robustness}

We further stress-test \model\ by perturbing the input root-space joints with zero-mean isotropic Gaussian noise of standard deviation
\(\sigma \in \{0, 2.5, 5, 10, 20, 40\}\) mm.
This probes how long the learned IK mapping remains usable as the observed skeleton departs from the articulated manifold seen during training.

\pgfplotstableread[row sep=\\]{
sigma_mm mpjae mpjpe p_mpjpe pve p_pve swing_mean twist_mean \\
0.0   7.4803 18.6120 12.7960 24.4794 18.0773 4.235 5.130 \\
2.5   7.6060 19.0023 13.1027 25.4350 18.6349 4.454 5.268 \\
5.0   7.9506 20.0860 13.9827 27.8565 20.1149 4.974 5.631 \\
10.0  9.1164 23.9635 17.0728 35.3516 24.9223 6.426 6.753 \\
20.0 12.4109 35.3930 26.0145 54.3116 37.5091 9.968 9.828 \\
40.0 20.2429 62.1124 46.8105 94.3768 63.3716 17.589 17.447 \\
}\noisedata

\begin{table}[htbp]
\centering
\small
\caption{\textbf{Robustness to Gaussian noise.} Aggregate metrics vs. noise level $\sigma$ (mm). Performance degrades sharply beyond $10$–$20$ mm.}
\label{tab:noise_robustness}
\pgfplotstabletypeset[
    columns={sigma_mm,mpjae,mpjpe,p_mpjpe,pve,p_pve},
    columns/sigma_mm/.style={column name={$\sigma$ (mm)}},
    columns/mpjae/.style={column name={MPJAE}},
    columns/mpjpe/.style={column name={MPJPE}},
    columns/p_mpjpe/.style={column name={P-MPJPE}},
    columns/pve/.style={column name={MPVE}},
    columns/p_pve/.style={column name={P-MPVE}},
    every head row/.style={before row=\hline,after row=\hline},
    every last row/.style={after row=\hline},
    precision=2,
    fixed,
    col sep=space
]{\noisedata}
\end{table}

\begin{figure}[htbp]
\centering
\begin{tikzpicture}
\begin{groupplot}[
    group style={group size=1 by 2, vertical sep=1.7cm},
    width=0.9\linewidth,
    height=3.4cm,
    xlabel={Noise standard deviation $\sigma$ (mm)},
    grid=major,
    tick label style={font=\scriptsize},
    label style={font=\scriptsize},
    legend style={font=\scriptsize, draw=none, fill=none, at={(0.5,0.95)}, anchor=south, legend columns=-1},
    xmin=0, xmax=40,
    xtick={0,2.5,5,10,20,40},
]

\nextgroupplot[
    ylabel={Rotation error},
    title={Rotation-space robustness},
]
\addplot+[mark=*] table[x=sigma_mm, y=mpjae] {\noisedata};
\addlegendentry{MPJAE}
\addplot+[mark=square*] table[x=sigma_mm, y=swing_mean] {\noisedata};
\addlegendentry{Mean swing}
\addplot+[mark=triangle*] table[x=sigma_mm, y=twist_mean] {\noisedata};
\addlegendentry{Mean twist}

\nextgroupplot[
    ylabel={Geometry error (mm)},
    title={Downstream geometry robustness},
]
\addplot+[mark=*] table[x=sigma_mm, y=mpjpe] {\noisedata};
\addlegendentry{MPJPE}
\addplot+[mark=square*] table[x=sigma_mm, y=p_mpjpe] {\noisedata};
\addlegendentry{P-MPJPE}
\addplot+[mark=triangle*] table[x=sigma_mm, y=pve] {\noisedata};
\addlegendentry{MPVE}
\addplot+[mark=diamond*] table[x=sigma_mm, y=p_pve] {\noisedata};
\addlegendentry{P-MPVE}

\end{groupplot}
\end{tikzpicture}
\caption{\textbf{Robustness curves.} Rotation (top) and geometry (bottom) errors vs. input noise. Stable at low noise, rapid degradation beyond plausible poses.}
\label{fig:noise_robustness_curves}
\end{figure}
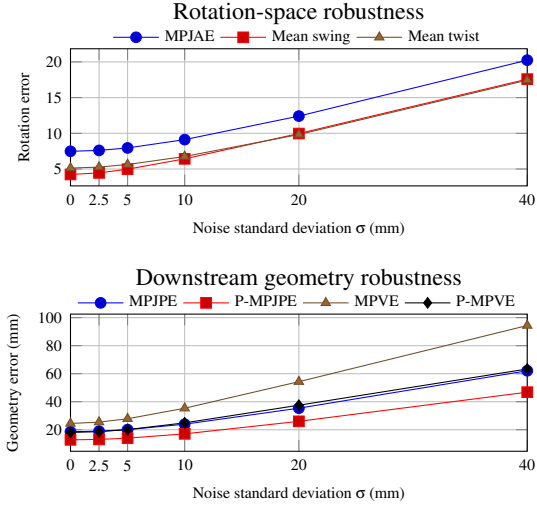

\pgfplotstableread[row sep=\\]{
x y value \\
0 0 5.028 \\ 1 0 2.284 \\ 2 0 2.352 \\ 3 0 2.821 \\ 4 0 1.937 \\ 5 0 2.375 \\ 6 0 3.437 \\ 7 0 1.586 \\ 8 0 1.977 \\ 9 0 2.472 \\ 10 0 3.104 \\ 11 0 2.735 \\ 12 0 2.907 \\ 13 0 4.914 \\ 14 0 5.406 \\ 15 0 6.930 \\ 16 0 3.027 \\ 17 0 2.527 \\ 18 0 2.436 \\ 19 0 2.744 \\ 20 0 14.277 \\ 21 0 15.886 \\
0 1 5.074 \\ 1 1 2.344 \\ 2 1 2.416 \\ 3 1 2.945 \\ 4 1 2.024 \\ 5 1 2.437 \\ 6 1 3.498 \\ 7 1 2.385 \\ 8 1 2.669 \\ 9 1 2.519 \\ 10 1 3.710 \\ 11 1 3.363 \\ 12 1 3.144 \\ 13 1 5.085 \\ 14 1 5.593 \\ 15 1 7.196 \\ 16 1 3.141 \\ 17 1 2.667 \\ 18 1 2.619 \\ 19 1 2.914 \\ 20 1 14.320 \\ 21 1 15.918 \\
0 2 5.201 \\ 1 2 2.534 \\ 2 2 2.583 \\ 3 2 3.262 \\ 4 2 2.242 \\ 5 2 2.618 \\ 6 2 3.679 \\ 7 2 3.894 \\ 8 2 4.069 \\ 9 2 2.675 \\ 10 2 5.031 \\ 11 2 4.676 \\ 12 2 3.842 \\ 13 2 5.595 \\ 14 2 6.133 \\ 15 2 7.963 \\ 16 2 3.446 \\ 17 2 3.003 \\ 18 2 3.124 \\ 19 2 3.385 \\ 20 2 14.434 \\ 21 2 16.040 \\
0 3 5.688 \\ 1 3 3.169 \\ 2 3 3.182 \\ 3 3 4.272 \\ 4 3 2.978 \\ 5 3 3.253 \\ 6 3 4.310 \\ 7 3 7.263 \\ 8 3 7.293 \\ 9 3 3.233 \\ 10 3 8.124 \\ 11 3 7.787 \\ 12 3 5.814 \\ 13 3 7.414 \\ 14 3 8.000 \\ 15 3 10.239 \\ 16 3 4.537 \\ 17 3 4.071 \\ 18 3 4.576 \\ 19 3 4.747 \\ 20 3 14.922 \\ 21 3 16.503 \\
0 4 7.225 \\ 1 4 4.834 \\ 2 4 4.806 \\ 3 4 6.943 \\ 4 4 4.878 \\ 5 4 4.958 \\ 6 4 6.352 \\ 7 4 14.265 \\ 8 4 14.068 \\ 9 4 4.925 \\ 10 4 14.764 \\ 11 4 14.461 \\ 12 4 10.254 \\ 13 4 12.247 \\ 14 4 12.824 \\ 15 4 15.526 \\ 16 4 7.515 \\ 17 4 6.869 \\ 18 4 8.134 \\ 19 4 8.248 \\ 20 4 17.016 \\ 21 4 18.187 \\
0 5 11.104 \\ 1 5 8.618 \\ 2 5 8.663 \\ 3 5 12.565 \\ 4 5 9.019 \\ 5 5 8.733 \\ 6 5 11.396 \\ 7 5 27.578 \\ 8 5 27.039 \\ 9 5 8.888 \\ 10 5 27.903 \\ 11 5 27.403 \\ 12 5 18.946 \\ 13 5 22.883 \\ 14 5 23.271 \\ 15 5 25.054 \\ 16 5 14.012 \\ 17 5 13.154 \\ 18 5 16.139 \\ 19 5 16.158 \\ 20 5 23.807 \\ 21 5 24.561 \\
}\swingjointdata

\pgfplotstableread[row sep=\\]{
x y value \\
0 0 1.739 \\ 1 0 2.842 \\ 2 0 2.814 \\ 3 0 1.802 \\ 4 0 3.935 \\ 5 0 4.224 \\ 6 0 1.230 \\ 7 0 3.886 \\ 8 0 3.746 \\ 9 0 1.451 \\ 10 0 3.753 \\ 11 0 4.074 \\ 12 0 4.665 \\ 13 0 3.505 \\ 14 0 3.071 \\ 15 0 7.517 \\ 16 0 3.662 \\ 17 0 4.573 \\ 18 0 12.099 \\ 19 0 7.299 \\ 20 0 19.496 \\ 21 0 11.488 \\
0 1 2.047 \\ 1 1 2.975 \\ 2 1 2.966 \\ 3 1 2.042 \\ 4 1 4.039 \\ 5 1 4.330 \\ 6 1 1.411 \\ 7 1 3.922 \\ 8 1 3.829 \\ 9 1 1.590 \\ 10 1 3.806 \\ 11 1 4.146 \\ 12 1 4.982 \\ 13 1 3.517 \\ 14 1 3.089 \\ 15 1 8.042 \\ 16 1 3.743 \\ 17 1 4.673 \\ 18 1 12.175 \\ 19 1 7.420 \\ 20 1 19.551 \\ 21 1 11.600 \\
0 2 2.758 \\ 1 2 3.370 \\ 2 2 3.405 \\ 3 2 2.660 \\ 4 2 4.384 \\ 5 2 4.656 \\ 6 2 1.802 \\ 7 2 4.038 \\ 8 2 4.042 \\ 9 2 1.940 \\ 10 2 3.965 \\ 11 2 4.345 \\ 12 2 5.731 \\ 13 2 3.541 \\ 14 2 3.155 \\ 15 2 9.424 \\ 16 2 4.008 \\ 17 2 4.926 \\ 18 2 12.367 \\ 19 2 7.771 \\ 20 2 19.676 \\ 21 2 11.916 \\
0 3 4.507 \\ 1 3 4.617 \\ 2 3 4.752 \\ 3 3 4.201 \\ 4 3 5.598 \\ 5 3 5.794 \\ 6 3 2.836 \\ 7 3 4.502 \\ 8 3 4.778 \\ 9 3 2.898 \\ 10 3 4.533 \\ 11 3 5.115 \\ 12 3 7.899 \\ 13 3 3.728 \\ 14 3 3.398 \\ 15 3 13.221 \\ 16 3 4.949 \\ 17 3 5.748 \\ 18 3 13.096 \\ 19 3 9.064 \\ 20 3 20.211 \\ 21 3 13.123 \\
0 4 8.566 \\ 1 4 7.881 \\ 2 4 8.179 \\ 3 4 7.854 \\ 4 4 8.943 \\ 5 4 8.873 \\ 6 4 5.225 \\ 7 4 6.110 \\ 8 4 7.022 \\ 9 4 5.217 \\ 10 4 6.352 \\ 11 4 7.598 \\ 12 4 12.665 \\ 13 4 4.692 \\ 14 4 4.445 \\ 15 4 20.929 \\ 16 4 7.760 \\ 17 4 8.369 \\ 18 4 16.081 \\ 19 4 13.284 \\ 20 4 23.001 \\ 21 4 17.170 \\
0 5 17.605 \\ 1 5 15.135 \\ 2 5 15.994 \\ 3 5 15.950 \\ 4 5 17.378 \\ 5 5 16.888 \\ 6 5 10.567 \\ 7 5 11.815 \\ 8 5 13.558 \\ 9 5 10.376 \\ 10 5 12.369 \\ 11 5 14.553 \\ 12 5 22.077 \\ 13 5 8.279 \\ 14 5 8.024 \\ 15 5 33.159 \\ 16 5 14.353 \\ 17 5 15.136 \\ 18 5 25.221 \\ 19 5 24.588 \\ 20 5 31.969 \\ 21 5 28.837 \\
}\twistjointdata

\begin{figure}[htbp]
\centering
\begin{tikzpicture}
\begin{groupplot}[
    group style={group size=1 by 2, vertical sep=1.2cm},
    width=0.9\linewidth,
    height=3.5cm,
    enlargelimits=false,
    ymin=-0.5, ymax=5.5,
    y dir=reverse,
    ytick={0,1,2,3,4,5},
    yticklabels={0,2.5,5,10,20,40},
    ylabel={Noise $\sigma$ (mm)},
    xtick={0,...,21},
    xticklabels={pel,lh,rh,s1,lk,rk,s2,la,ra,s3,lf,rf,n,lc,rc,h,ls,rs,le,re,lw,rw},
    x tick label style={rotate=90,anchor=east,font=\scriptsize},
    y tick label style={xshift=-5pt},
    tick label style={font=\scriptsize},
    label style={font=\scriptsize},
    colorbar,
    point meta min=0,
    point meta max=35,
    colormap/viridis,
]

\nextgroupplot[
    title={Per-joint swing error},
    colorbar style={title={deg}, yticklabel style={font=\scriptsize}, width=10pt},
]
\addplot[
    scatter,
    only marks,
    mark=square*,
    mark size=4.25pt,
    scatter src=explicit,
] table[x=x, y=y, meta=value] {\swingjointdata};

\nextgroupplot[
    title={Per-joint twist error},
    colorbar style={title={deg}, yticklabel style={font=\scriptsize}, width=10pt},
]
\addplot[
    scatter,
    only marks,
    mark=square*,
    mark size=4.25pt,
    scatter src=explicit,
] table[x=x, y=y, meta=value] {\twistjointdata};

\end{groupplot}
\end{tikzpicture}
\caption{\textbf{Per-joint degradation under noise.} Swing (top) and twist (bottom) error across joints and noise levels. Distal joints fail first; proximal joints remain more stable.}
\label{fig:noise_per_joint_heatmaps}
\end{figure}
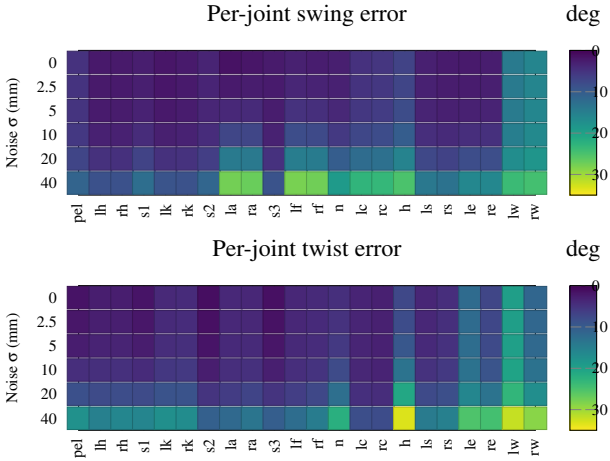

The results in Tab.~\ref{tab:noise_robustness} and Fig.~\ref{fig:noise_robustness_curves} show two clear trends. First, the model is stable under mild perturbations: up to \(5\) mm noise, MPJAE increases only from \(7.48^\circ\) to \(7.95^\circ\), and P-MPJPE from \(12.80\) mm to \(13.98\) mm. Second, degradation accelerates beyond \(10\) mm and becomes severe at \(20\) mm and \(40\) mm, where the input skeleton is no longer a plausible tracked pose. Figure~\ref{fig:noise_per_joint_heatmaps} shows that failure starts in distal, swing-dominated joints such as ankles and feet, while pelvis and spine remain comparatively stable. Twist degrades more gradually at moderate noise, consistent with Gaussian perturbations first corrupting local bone directions before fully destabilizing the rotational frame.

\subsection{Per-joint rotation error breakdown}
\label{app:per_joint_breakdown}

Tab~\ref{tab:per_joint_metrics_appendix} provides the exact values underlying Fig~\ref{fig:per_joint_improvement}. The largest gains appear on distal lower-limb joints, especially ankles and feet, where IK-GAT reduces MPJAE by roughly half relative to the Transformer. Since these joints are highly sensitive to accumulated upstream error, this supports the claim that bidirectional kinematic message passing helps stabilize long chains instead of predicting each joint in isolation.

\begin{table}[t]
\small
\setlength{\tabcolsep}{4pt}
\caption{\textbf{Per-joint rotation error on AMASS.} IK-GAT vs. Transformer. MPJAE with swing and twist components per joint.}
\label{tab:per_joint_metrics_appendix}
\begin{tabularx}{\linewidth}{|l|YYY|YYY|}
\hline
& \multicolumn{3}{c|}{\textbf{Transformer}} & \multicolumn{3}{c|}{\textbf{IK-GAT (ours)}} \\
\hline
\textbf{Joint} & \textbf{MPJAE} & \textbf{Swing} & \textbf{Twist} & \textbf{MPJAE} & \textbf{Swing} & \textbf{Twist} \\
\hline
pelvis         & 5.883 & 3.996 & 3.670 & 3.956 & 3.105 & 1.989 \\
left\_hip      & 5.902 & 3.420 & 4.199 & 4.740 & 3.043 & 3.050 \\
right\_hip     & 5.884 & 3.635 & 3.995 & 4.339 & 2.813 & 2.813 \\
spine1         & 6.515 & 4.726 & 3.706 & 3.864 & 2.864 & 2.151 \\
left\_knee     & 7.221 & 3.220 & 5.947 & 5.256 & 2.714 & 3.995 \\
right\_knee    & 6.834 & 3.171 & 5.533 & 5.621 & 2.935 & 4.246 \\
spine2         & 6.458 & 5.490 & 2.699 & 4.864 & 4.200 & 1.924 \\
left\_ankle    & 8.939 & 7.670 & 3.736 & 3.983 & 2.024 & 3.091 \\
right\_ankle   & 9.231 & 7.115 & 4.871 & 4.632 & 2.311 & 3.623 \\
spine3         & 5.062 & 3.844 & 2.668 & 3.595 & 2.680 & 2.013 \\
left\_foot     & 8.958 & 7.713 & 3.676 & 3.956 & 1.916 & 3.137 \\
right\_foot    & 9.082 & 6.972 & 4.817 & 4.842 & 2.345 & 3.849 \\
neck           & 10.406 & 6.346 & 7.087 & 7.417 & 2.635 & 6.463 \\
left\_collar   & 7.782 & 6.313 & 3.628 & 5.133 & 3.621 & 3.060 \\
right\_collar  & 8.983 & 7.608 & 3.650 & 5.240 & 3.699 & 3.085 \\
head           & 20.117 & 13.216 & 12.960 & 16.512 & 9.544 & 11.819 \\
left\_shoulder & 8.132 & 5.167 & 5.372 & 6.039 & 3.054 & 4.679 \\
right\_shoulder& 7.986 & 4.622 & 5.702 & 5.751 & 3.102 & 4.297 \\
left\_elbow    & 14.510 & 6.235 & 12.168 & 11.882 & 2.852 & 11.132 \\
right\_elbow   & 16.932 & 5.792 & 15.024 & 12.716 & 2.595 & 12.111 \\
left\_wrist    & 30.132 & 18.432 & 21.110 & 29.401 & 16.869 & 21.376 \\
right\_wrist   & 31.257 & 17.298 & 23.284 & 26.934 & 14.303 & 20.387 \\
\hline
\textbf{Mean}  & 11.009 & 6.909 & 7.250 & 8.212 & 4.328 & 6.104 \\
\textbf{Std}   & 7.199  & 4.090 & 5.748 & 7.084 & 3.873 & 5.556 \\
\hline
\end{tabularx}
\end{table}

Many of these gains come primarily from \emph{swing} rather than \emph{twist}, especially at the ankles, feet, neck, and elbows. This is consistent with the proposed bone-aligned representation: it simplifies the geometric component of the problem by aligning the local frame with the bone direction, while leaving the intrinsic ambiguity of twist from positions alone largely unresolved.
The hardest joints remain the head, elbows, and especially wrists. For wrists, the absolute error stays high for both models, which is expected because they are leaf joints with very limited positional evidence for axial rotation. By contrast, proximal joints such as pelvis, hips, and spine show smaller but consistent gains, indicating that IK-GAT improves the rotation field across the whole body while yielding the largest benefits where inverse kinematics is most ambiguous.

\pgfplotstableread[col sep=comma]{
batch_size,cpu_fps,cuda_fps
1,693.30,447.96
2,1223.51,901.45
4,1925.73,1804.00
8,2898.38,3640.07
16,4446.81,7201.70
32,5849.37,14244.76
64,7638.05,28018.18
}\runtimefpsdata

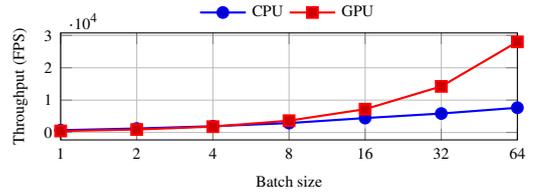
\begin{figure}[t]
\centering
\begin{tikzpicture}
\begin{axis}[
    width=0.9\linewidth,
    height=3cm,
    xmode=log,
    log basis x={2},
    xmin=1, xmax=64,
    xtick={1,2,4,8,16,32,64},
    xticklabels={1,2,4,8,16,32,64},
    xlabel={Batch size},
    ylabel={Throughput (FPS)},
    grid=major,
    tick label style={font=\scriptsize},
    label style={font=\scriptsize},
    legend style={
        font=\scriptsize,
        draw=none,
        at={(0.5,1.02)},
        anchor=south,
        legend columns=2,
    },
]
\addplot+[mark=*, thick] table[x=batch_size, y=cpu_fps] {\runtimefpsdata};
\addlegendentry{CPU}
\addplot+[mark=square*, thick] table[x=batch_size, y=cuda_fps] {\runtimefpsdata};
\addlegendentry{GPU}
\end{axis}
\end{tikzpicture}
\caption{\textbf{Inference throughput vs. batch size.} CPU is faster at $B=1$, while GPU scales better and dominates for larger batches.}
\label{fig:runtime_batchsize}
\end{figure}

\begin{figure*}[htbp]
    \centering
    \includegraphics[width=0.925\linewidth]{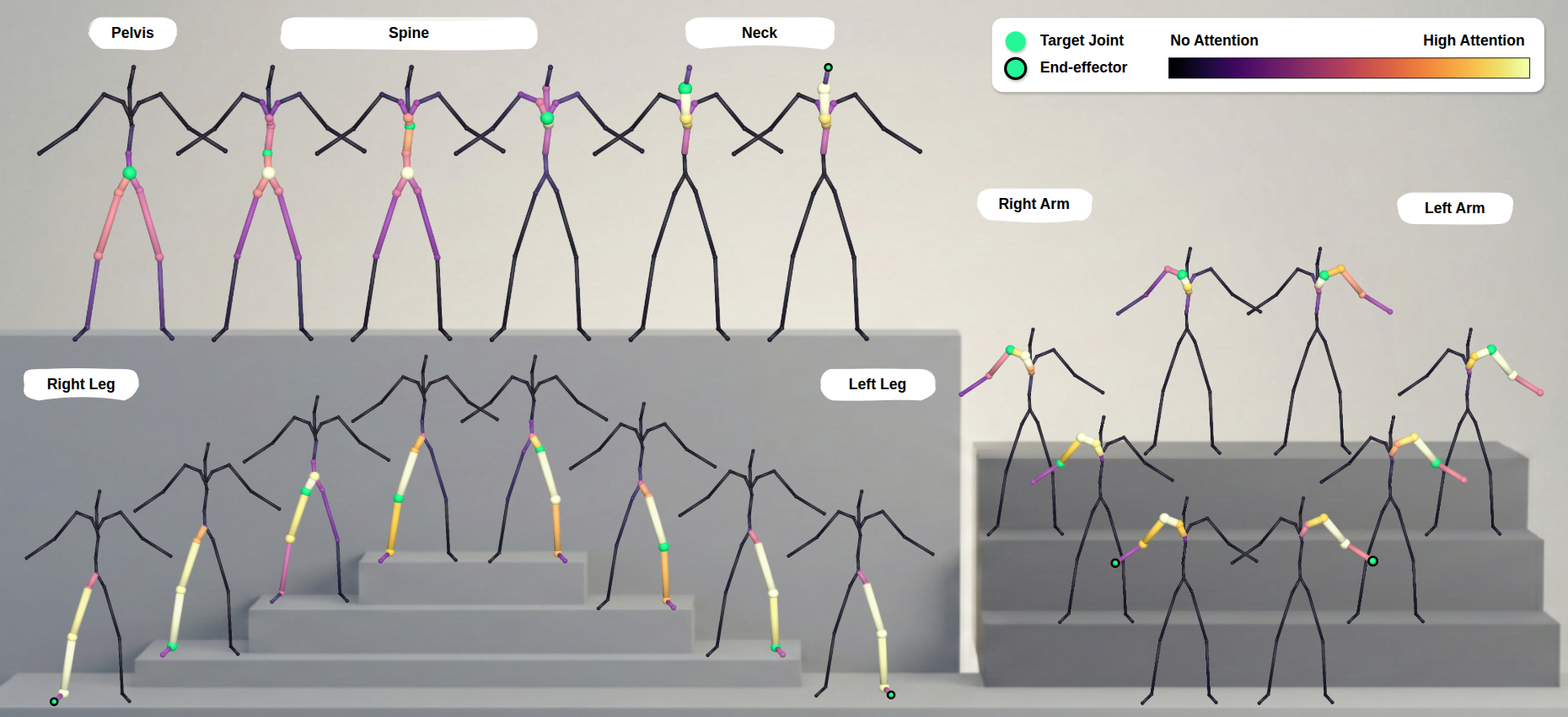}
    \caption{\textbf{Effective multi-layer attention flow on learned kinematic graph} follows anatomically meaningful pathways: pelvis and torso act as coordination hubs, proximal joints dominate distal predictions, and symmetric limbs exhibit broadly mirrored context patterns.}
    \label{fig:attention-flow-smpl}
\end{figure*}

\subsection{Inference runtime analysis}
\label{app:runtime_analysis}

We evaluate the throughput of the pure IK inference path (network forward pass + analytic local rotation recovery) across batch sizes on AMD Ryzen 9 7900X3D 12-Core CPU and NVIDIA GeForce RTX4060 Ti GPU in Fig~\ref{fig:runtime_batchsize}.
At \(B=1\), CPU is faster than GPU (\(693\) FPS vs.~\(448\) FPS), indicating that single-frame inference is dominated by fixed overheads rather than compute. The gap closes quickly, and the crossover occurs between \(B=4\) and \(B=8\). Beyond that point, GPU throughput scales much better, reaching \(3640\) FPS at \(B=8\), \(7202\) FPS at \(B=16\), and \(28018\) FPS at \(B=64\), compared to \(2898\), \(4447\), and \(7638\) FPS on CPU. This separates two deployment regimes: CPU is favorable for strictly online inference, while GPU is clearly preferable for batched processing.

\subsection{Attention Flow}

Figure~\ref{fig:attention-flow-smpl} visualizes the effective multi-layer attention flow of \model\ for individual target joints on the SMPL skeleton. For each target joint, highlighted in green, attention is propagated backward through the stacked graph-attention layers to estimate how strongly each joint influences the final prediction. The figure is best read together with the graph-structure ablation in the main paper, since it helps explain why the bidirectional graph works: information is propagated along anatomically plausible routes rather than through unrestricted global interaction.
The attention flow follows clear kinematic structure. Axial joints aggregate information mainly along the torso while also integrating bilateral context from shoulders and pelvis. Limb joints are dominated by chain-local influence, with stronger dependence on proximal than distal nodes. The pelvis acts as a global coordination hub across both legs and the spine. End-effectors show reduced self-influence and depend more strongly on upstream joints, which is consistent with inverse kinematics, where distal orientations are largely constrained by proximal pose. Overall, the patterns indicate that \model\ learns hierarchical, skeleton-aware information propagation rather than independent per-joint regression.

\section{Baseline Architectures}
\label{app:baseline-architectures}

This appendix provides architecture details for the in-house baselines used to isolate the effect of inductive bias in inverse kinematics. All models take root-space joint positions as input, predict per-joint rotations in the same 6D representation as \model, and are trained with the same loss and protocol unless stated otherwise.

\noindent
\textbf{MLP (per-joint regression baseline).}
The MLP baseline treats inverse kinematics as independent per-token regression. Each joint is projected to a latent space, augmented with learnable joint and temporal embeddings, processed by residual feed-forward blocks, and mapped to per-joint rotations with a linear head. Since there is no explicit interaction between joints, this baseline tests whether position-to-rotation mapping can be solved as a purely local regression problem.

\noindent
\textbf{Spatio-Temporal IK (STIK).}
STIK introduces both kinematic structure and temporal modeling. Per-joint point features are combined with parent-conditioned features, processed with an LSTM and self-attention, and then aggregated with a causal temporal convolutional network. This makes it a stronger baseline than the MLP, but its spatial structure is injected indirectly through feature design and sequential processing rather than explicit graph message passing.

\noindent
\textbf{Transformer (fully-connected attention baseline).}
The Transformer baseline applies global self-attention over flattened joint-time tokens after input projection and joint/temporal embeddings. It can model unrestricted global dependencies, but it has no explicit notion of kinematic adjacency. This baseline tests whether generic dense attention alone is sufficient for inverse kinematics.

\noindent
\textbf{Adaptation of QuaterGCN.}
In its original form, QuaterGCN~\cite{song2024quatergcn} takes 2D joint positions and derived 2D angles as input and predicts quaternions representing 3D joint rotations. We adapt it to take 3D joint positions and evaluate both the original quaternion output and a 6D continuous rotation output.

\end{document}